\documentclass[a4paper,fleqn]{cas-dc}

\usepackage{natbib}
\usepackage{bigstrut}
\usepackage{pifont}
\usepackage{subfigure}
\usepackage{hyperref}       
\usepackage{url}            
\usepackage{amsfonts}       
\usepackage{nicefrac}       
\usepackage{microtype}      
\usepackage{xcolor}         
\usepackage{amssymb}
\usepackage{graphicx}
\usepackage{tabularx}
\usepackage{multirow}
\usepackage{paralist}
\usepackage{enumitem}
\usepackage{booktabs}
\usepackage{caption}
\usepackage[switch]{lineno}
\usepackage{siunitx}

\def\tsc#1{\csdef{#1}{\textsc{\lowercase{#1}}\xspace}}
\tsc{WGM}
\tsc{QE}
\tsc{EP}
\tsc{PMS}
\tsc{BEC}
\tsc{DE}

\newcommand{\subsubsubsection}[1]{\paragraph{#1}\mbox{}\\} 
\begin{document} 
\begin{sloppypar}
\let\WriteBookmarks\relax
\def\floatpagepagefraction{1}
\def\textpagefraction{.001}
\shorttitle{Cloud Removal With PolSAR-Optical Data Fusion Using A Two-Flow Residual Network}
\shortauthors{Y. Wang, et~al.}

\title [mode = title]{Cloud Removal With PolSAR-Optical Data Fusion Using A Two-Flow Residual Network}                      

\author[1,2,3]{Yuxi Wang}[style=chinese]
\ead{wangyuxi20@mails.ucas.ac.cn}
\address[1]{Key Laboratory of Digital Earth Science, Aerospace Information Research Institute, Chinese Academy of 
Sciences, Beijing 100094, China}
\address[2]{International Research Center of Big Data for Sustainable Development Goals, Beijing 100094, China}
\address[3]{College of Resources and Environment, University of Chinese Academy of Sciences, Beijing 100049, China}

\author[4]{Wenjuan Zhang}[style=chinese]
\ead{zhangwj@aircas.ac.cn}
\address[4]{Aerospace Information Research Institute, Chinese Academy of Sciences, Beijing 100094, China}
\cormark[1]
\cortext[cor1]{Corresponding author}

\author[1,2,3]{Bing Zhang}[style=chinese]
\ead{zb@radi.ac.cn}

\begin{abstract}
    Optical remote sensing images play a crucial role in the observation of the Earth's surface. However, obtaining complete optical remote sensing images is challenging due to cloud cover. Reconstructing cloud-free optical images has become a major task in recent years. This paper presents a two-flow Polarimetric Synthetic Aperture Radar (PolSAR)-Optical data fusion cloud removal algorithm (PODF-CR), which achieves the reconstruction of missing optical images. PODF-CR consists of an encoding module and a decoding module. The encoding module includes two parallel branches that extract PolSAR image features and optical image features. To address speckle noise in PolSAR images, we introduce dynamic filters in the PolSAR branch for image denoising. To better facilitate the fusion between multimodal optical images and PolSAR images, we propose fusion blocks based on cross-skip connections to enable interaction of multimodal data information. The obtained fusion features are refined through an attention mechanism to provide better conditions for the subsequent decoding of the fused images. In the decoding module, multi-scale convolution is introduced to obtain multi-scale information. Additionally, to better utilize comprehensive scattering information and polarization characteristics to assist in the restoration of optical images, we use a dataset for cloud restoration called OPT-BCFSAR-PFSAR, which includes backscatter coefficient feature images and polarization feature images obtained from PoLSAR data and optical images. Experimental results demonstrate that this method outperforms existing methods in both qualitative and quantitative evaluations.

\end{abstract}


\maketitle

\section{Introduction}
\label{sec:intro}
\subsection{Motivation}

Optical satellite images have played an increasingly important role in Earth observation in recent years. However, optical remote sensing satellites are inevitably affected by cloud occlusion when they acquire ground information. An examination of MODIS data indicates that 55\% of the land is covered with clouds \citep{ref1}. Cloud coverage leads to data loss, which prevents the obtained remote sensing image data from meeting the requirements for research and applications, particularly in quantitative remote sensing and the production of high-quality surface products in large areas. Therefore, it is of great value to reconstruct the missing areas caused by cloud cover.
Generally speaking, cloud removal is a process of missing information reconstruction which leverages complementary information to reconstruct the missed remote sensing image data contaminated by clouds. According to the different types of auxiliary information, the existing reconstruction methods can be divided into the following two categories: optical-based cloud removal methods and synthetic aperture radar (SAR)-based cloud removal methods \citep{ref2}. The optical-based cloud removal methods use the information of the missing optical image itself or the auxiliary information provided by other optical images to reconstruct the missing information in the image. It mainly includes spatial-based methods and temporal-based methods.

The spatial-based methods are derived from the natural image restoration. Based on the spatial correlation between local or non-local regions of the missing images, these methods estimate the pixels in missing regions with information obtained from undamaged regions in the same remote sensing image without the assistance of other information \citep{ref3}. It mainly includes variational methods \citep{ref4, ref5}, spatial interpolation methods \citep{ref6, ref7}, partial differential equation (PDE)-based methods \citep{ref8}, exemplar-based methods \citep{ref9, ref10}, and learning-based method \citep{ref11}. For example, Pan et al. \citep{ref11} proposed a spatial attention generative adversarial network (SpAGAN), which focuses the cloud area with local-to-global spatial attention, thereby enhancing the information of missing areas and generating high-quality cloud-free images. 

In general, the spatial-based methods are qualified for reconstructing small missing areas or regions with regular textures. However, for the heterogeneous texture area with large area thick cloud cover or complex surface object type, the reconstruction precision cannot be guaranteed due to the lack of additional auxiliary information.
The temporal-based methods are proposed to address the problem of large missing areas, using satellite imagery of the same area obtained at adjacent times as an auxiliary source. Based on the temporal correlation of the images, a functional transfer relationship between the cloudless data and the missing data is established to achieve information reconstruction \citep{ref12}. Compared with the spatial-based methods, the temporal-based methods are less affected by the size of cloud-covered areas and texture heterogeneity and can work well for cloud removal. However, when significant changes occur in a short period of time or cloudless reference data from adjacent time periods cannot be obtained, these methods may produce large reconstruction errors.

In recent years, a series of works have explored cloud removal methods using synthetic aperture radar (SAR) images as auxiliary data. SAR images are not affected by clouds due to its high penetration of cloud and cloud shadow, thus largely mitigating the challenge of cloud removal. The SAR-based cloud removal methods use SAR images as auxiliary information to repair missing areas on optical images, which effectively avoids the problem of missing region caused by clouds in the optical auxiliary images.

Single-polarization SAR images can only transmit and receive electromagnetic waves in a single direction, providing limited information about objects. With the development of SAR technology, polarimetric synthetic aperture radar (PolSAR) has become a new research focus in the field of microwave imaging systems \citep{ref13}. Compared with single-polarization SAR, PolSAR adopts alternating transmission and simultaneous reception to obtain four types of scattered echo signals, allowing for the recording the polarimetric scattering characteristics of different polarization channels of targets, which can be used to analyze the scattering mechanism \citep{ref14}.

Eckardt et al. \citep{ref15} first used multi-frequency SAR data as a reference and employed a geographically weighted model to remove clouds. With the development of deep learning, some studies have attempted to directly generate SAR images from optical images \citep{ref16, ref17, ref18}. For example, Bermudez et al. \citep{ref17} directly applied a conditional generative adversarial (cGAN) network to transform Sentinel-1 SAR data into optical images. However, these methods only use SAR data for restoration, causing changes in undamaged areas as well. Additionally, the optical images are only utilized as ground truth in the loss function during model training, and the information from undamaged regions in the optical images is generally not utilized.

Also, some studies have studied the comprehensive application of both cloudy optical images and their corresponding SAR images to delve into the characteristics of the two imaging modes. Grohnfeldt et al. \citep{ref19} jointly used the C-band VV polarization SAR data and damaged optical images based on a SAR-Optical-cGAN network (SAR-Opt-cGAN) to restore missing areas in Sentinel-2 images, effectively removing thin cloud areas. Gao et al. \citep{ref20} used a U-net simulation network to simulate optical images from SAR images, and further utilized a fusion cGAN network to combine damaged optical data, VV and VH polarization SAR data, and simulated optical data to obtain cloud-free outputs with high-frequency textures, achieving good reconstruction results in agricultural fields, forests, and roads. Considering that cGAN may exhibit training and prediction instability in the case of large-scale missing data, Meraner et al. \citep{ref21} proposed a deep residual convolutional neural network (DSen2-CR) to reconstruct Sentinel-2 images with the assistance of VV and VH polarization Sentinel-1 data. Experimental validation on a public large-scale dataset (i.e., SEN12MS-CR) confirmed the robustness to thick clouds in optical images and the model's applicability to various scenarios. Considering that convolutional layers cannot capture global contextual information and non-local features cannot be effectively represented, recent research has introduced attention mechanisms to improve models, enabling the establishment of non-local relationship models and utilizing global information for subsequent restoration.

Han et al. \citep{ref22} proposed a cloud removal method based on the transformer, which effectively extracts global contextual information from VV, VH-polarized SAR images, and optical images using self-attention mechanisms, generating high-quality cloud-free images with high global consistency. In addition, they designed an image preprocessor (IPP) and a denoising image restorer with channel attention mechanism (decloud-IR) to handle multi-channel inputs and outputs images, fully considering the contributions of different channels in the feature maps to generate remote sensing images. Wang et al. \citep{ref23} advanced a unified spatial-spectral residual network for cloud removal (USSRN-CR), which utilizes VV and VH-polarized SAR images as auxiliary data to remove clouds from optical images. They introduced gated convolutional layers to distinguish cloud pixels from clean pixels, preventing the influence of cloud area information on subsequent feature extraction. Also, they designed self-attention mechanism to obtain global spatial and spectral information, enabling better image restoration.
However, reconstruction of high-quality cloudless images based on SAR images is still a challenging due to the following problems: 

(1)SAR images usually have a large amount of speckle noise, which is generated by the coherent interference between by the radar echo signal and the target scatterer. Therefore, random fading of amplitude and phase will occur in the pixels, and intersecting speckle noise will appear in SAR images, which make it difficult to obtain the effective information in SAR images \citep{ref24}. 

(2)SAR images and optical images have different characteristic differences due to different imaging mechanisms \citep{ref25, ref26}. Therefore, it is crucial to fully realize the fusion of the multi-modal SAR images and optical images to reconstruct cloudy areas. 

Based on these experiences, most of the existing methods directly feed the single-polarization or dual-polarization SAR images and optical images into the network through a simple stack \citep{ref19, ref20, ref21, ref22, ref23}. However, single-polarization or dual-polarization data cannot obtain a comprehensive scattering information of ground objects, and the lack of the information of the polarization characteristics, which affects the accuracy of subsequent restoration. In addition, without the deep fusion of optical images and SAR images, the complementary information on SAR image cannot realize effectively cross-modality interaction. Finally, the fused images obtained by simply concatenation between multi-modality images are susceptible to speckle noise, which leads to the generation of cloud removal results with fuzzy details. 

With the aforementioned considerations, we fully consider the imaging characteristics of SAR images, and propose a two-flow PolSAR-Optical data fusion-based cloud removal (PODF-CR) method to reconstruct the contaminated regions. In order to better utilize comprehensive scattering information of ground objects, we used a dataset which includes PolSAR images and optical images for cloud removal called OPT- BCFSAR -PFSAR. Three optimization terms are constructed to constrain the model and the structural details in cloud removal results are enhanced. Specifically, PODF-CR contains two parallel backbones developed for optical and PolSAR image representation learning. In order to better realize the fusion between multi-modality optical images and PolSAR images, we have proposed the multi-modality cross fusion (MMCF) block and multi-modality refinement fusion (MMRF) block, that can capture interactions between multi-modality data and refine the extraction features. Considering the speckle noise and geometric distortion in PolSAR images, we introduce the coupling spatial and channel dynamic filters (SCDF) block to denoise SAR images for the fusion image decoding. 
To sum up, the main contributions of the proposed approach are summarized below:
\begin{itemize}
\item	We propose a novel two-flow cloud removal algorithm based on PolSAR-Optical data fusion named PODF-CR. We use encoding–decoding blocks to synergistically accomplish cloud removal.
\item	MMCF block and MMRF block are proposed for effective merging of multimodal PolSAR and optical data. MMCF block can accomplish fusion in shallow features and subsequently deep features, and it can more effectively realize the interaction between PolSAR image information and optical image information based on the cross-modality skip connection. MMRF block achieves the suppression of redundant and irrelevant information, highlighting of complementary information, and refinement of multi-modality features based on the attention mechanism. 
\item	In order to reduce the influence of coherent speckle noise on the generated cloudless optical images, SCDF module based on the spatial and channel dynamic filters are introduced to generate more reliable texture details. 
\item	To verify the feasibility and applicability of the method, we conducted experiments using the OPT- FPBCSAR -PFSAR dataset, which includes optical and PolSAR data. The PolSAR images as auxiliary information for restoration are introduced to describe the polarization characteristics of the target such as phase and reflect the comprehensive scattering information. The impact of missing images from different categories and with different cloud coverage levels on the proposed model was evaluated.
\end{itemize}

\subsection{Paper Structure}

The rest of this article is organized as follows. Section 2 briefly reviews the related work. In Section 3, we elaborate on the proposed model in detail. Section 4 presents experimental results and discussions on the established dataset. The conclusion is summarized in Section 5.


\section{Related Work}
\label{Related Work}
\subsection{Application of PolSAR Data}

Single polarized datas can only transmit and receive electromagnetic waves with fixed polarization techniques, so it cannot describe the phase and polarization properties of targets and is unable to gather comprehensive ground object scattering information \citep{ref27}. Nowadays, the SAR system is always evolving toward multifunctional, multiresolution, and multiworking versions. By sending and receiving various combinations of polarized waves, it may generate PolSAR data.

In contrast to single-polarization SAR, PolSAR data contains rich character, highly sensitive to the height, form, and direction as well as the spatial distribution of ground objects, and can also extract the entire polarization matrix and geometric structure details of targets \citep{ref28}. Several studies demonstrate that PolSAR data is more advantageous than single-polarization and dual-polarization data in target recognition and image classification \citep{ref29, ref30, ref31}. 

In recent years, the classification techniques for PolSAR images have been developed. The polarimetric decomposition approachs are the common method and utilize the polarization scattering information to improve the classification accuracy \citep{ref32}. These techniques extract different scattering features from the coherence matrix or covariance matrix of PolSAR images, such as Pauli decomposition \citep{ref33}, Freeman–Durden decomposition \citep{ref34}, Cloude–Pottier decomposition \citep{ref35}, Yamaguchi four-component decomposition \citep{ref36, ref37}, and so on. With the advancement of machine learning, some classifiers such as support vector machine (SVM) \citep{ref38} and K-nearest neighbor (KNN) \citep{ref39} have been used for PolSAR classification, and they performed better than polarimetric decomposition methods. Considering that machine learning algorithms need to manually extract features, numerous studies are using deep learning algorithms for classification which can fully mine data information.

For example, Zhou et al. \citep{ref40} applied convolutional networks to PolSAR image classification. They first represented the full-polarization data in the form of a coherence matrix, and then transformed the amplitude elements into standardized 6-dimensional real feature vectors, which were fed into a four-layer convolutional neural network (CNN). On Flevoland dataset, this neural network achieves 92\% accuracy. However, these studies directly apply neural networks to the amplitude of SAR images and ignore the phase information. However, the phase of the non-diagonal elements of the covariance/coherence matrix is highly helpful in identifying various scatterers for PolSAR data. Therefore, Zhang et al. \citep{ref41} introduced a complex-valued CNN (CV-CNN) specifically for SAR image interpretation, which simultaneously utilizes the amplitude and phase information. All layers of the network are extended to the complex domain. Fang et al. \citep{ref42} proposed a hybrid-attention-based encoder-decoder fully convolutional network (HA-EDNet) for PolSAR classification. They represented the real and imaginary parts of the diagonal and non-diagonal elements of the polarimetric coherence matrix as nine-dimensional vectors, using a self-attention module to establish global spatial dependencies and extract contextual features, further improving classification accuracy. On the Flevoland-15 dataset, the OA index reached 99\%.

In addition, there are also many studies on image classification based on the fusion of PolSAR data and optical data. Kussul \citep{ref43} classified crops (such as corn and soybeans) in the Ukrainian region using a combination of Radarsat-2 backscatter coefficient SAR images and EO-1/ALI optical images, comparing the classification accuracy of neural networks, SVM, and decision trees, with the neural network achieving the best classification accuracy of 80.4\%. La et al. \citep{ref44} based their study on PALSAR-2 SAR data and Sentinel-2 optical data. They represented the PolSAR data as the intensity data of four polarizations, entropy (H), alpha ($\alpha$), and anisotropy (A) data obtained based on Cloude decomposition, Yamaguchi's four components of volume scattering (Vol), double scattering (Dbl), odd scattering (Odd), and helix scattering (Hlx), and nine-dimensional vector data based on the coherence matrix. The experimental results showed that adding PALSAR-2 images improved the classification accuracy to 72.8\%. Therefore, compared to using only multispectral data, adding PolSAR data greatly improves the classification accuracy. Thus, the fusion of PolSAR images with optical images can provide more information about the land features.

Based on the existing research foundation, it can be observed that the current fusion of PolSAR data with optical images mainly involves two approaches. The first approach utilizes the backscatter characteristics of SAR images for fusion, while the second approach involves transforming full-polarization data into polarization matrices, such as the polarimetric coherence matrix. Based on the coherence matrix or the transformed matrix, polarization features obtained from polarization decomposition are used to achieve SAR and optical image fusion for subsequent research applications.

Therefore, compared to single-polarization or dual-polarization data, the joint application of the coherence matrix of PolSAR images with optical images can provide more scattering information and polarization characteristics of ground objects in the field of image restoration, which helps to analyze the texture details of missing image areas and improve the restoration accuracy of optical images. Nevertheless, existing cloud removal methods based on SAR images mostly simply concatenate single-polarization or dual-polarization SAR data with optical data as input to the network. In this paper, we represent the PolSAR data as polarization feature data and input it into the network for training.

\subsection{Multi-modality Data Fusion}

The primary technique used in simple multimodal fusion approaches is element-wise addition or multiplication operations. There are also methods that involve simply concatenating the multimodal data or their features into a deep neural network \citep{ref45, ref46}. Based on the position of the fusion, it can be classified as early fusion, middle fusion, and late fusion, etc. For instance, most algorithms for optical image restoration concatenate SAR images with cloudy optical images and input them into the network for subsequent restoration based on the early fusion, without effectively utilizing supplementary information \citep{ref19, ref20, ref21}. Concatenate-based fusion is widely used and has achieved success in feature extraction and representation, it still has limitations in utilizing multimodal data, resulting in limited performance gains \citep{ref47}. In order to better utilize the supplementary information from auxiliary data, Hong et al. \citep{ref48} developed multi-modality data fusion, aiming to learn more compact multi-modality feature representations by interactively modifying the parameters of different subnetworks. A network flow of one modality takes into account a wider range of supplements from another flow in addition to learning specific features from itself to achieve more comprehensive information integration. Xu et al. \citep{ref49} proposed a new global-local fusion-based cloud removal (GLF-CR) algorithm to obtain global texture information, which combines cross-modality fusion with a spatial self-attention mechanism. The algorithm uses complementary information embedded in SAR images to guide the interaction of global and local information, producing more reliable local texture details and ensuring consistency between the recovered areas and the overall areas. These methods have improved the exploitation of supplemental information from auxiliary data. In order to comprehensively and deeply integrate multimodal information, we have introduced MMCF and MMRF into the network. In the MMCF, we learn the advantages and complementarity of different modalities through skip connections, achieving better integration of multimodal data. In terms of MMRF, we consider the redundancy of feature information and global context information, refining features based on spatial-channel attention unit (SCAU)and multi-modality weighted refinement unit (MWRU).

\subsection{Dynamic filters}

Standard convolutional filters share the same weights across all pixels in an image, while dynamic filters vary their weights according to the input features, allowing for adaptive and flexible feature embedding. Jia et al. \citep{ref50} initially proposed a dynamic filter network, consisting of a filter generation network and a dynamic filtering layer. The former predicts the kernel based on the input image, and the latter applies the generated kernel to another input.

Dynamic filtering has made rapid progress in image denoising tasks in recent years. Niklaus et al. \citep{ref51} applied dynamic filters to video frame interpolation, using deep convolutional networks to estimate adaptive convolutional kernels for each pixel. This approach integrated optical flow estimation and frame synthesis into a framework, effectively suppressing problems including blurring, occlusion, and abrupt changes in brightness, resulting in high-quality video frames. In \citep{ref52}, Mildenhall et al. developed a dynamic filtering structure to predict spatially variable kernels, enabling image registration and denoising based on the kernel at each pixel. However, the predicted filters in the aforementioned techniques are the identical for each pixel across different channels and cannot encode channel-specific information. To handle spatially varying tasks more flexibly, Zhou et al. \citep{ref53} proposed a spatio-temporal filter adaptive network (STFAN) for alignment and deblurring within a unified framework. By using filter adaptive convolution (FAC) filters, the previous frame's deblurred features are aligned with the current frame, removing spatially varying blurriness from the current frame's features. However, this fully dynamic filtering consumes significant computational and memory costs and can only replace a few standard convolutional layers in a CNN. On this basis, Zhou et al. \citep{ref54} introduced decoupled dynamic filters (DDF), separating dynamic filters into spatial and channel dynamic filters, significantly reducing the number of parameters. Moreover, replacing standard convolutions with DDF also led to better classification accuracy.

Inspired by the use of dynamic filters for denoising, we introduce the SCDF which decouples spatial and channel dynamic filters into the network to achieve denoising of SAR images.

\section{Materials and Methodology}
\label{Materials and Methodology}
\subsection{Overview of the Cloud Removal Network}

\begin{figure*}[!t]
  \centering
  \includegraphics[width=\textwidth]{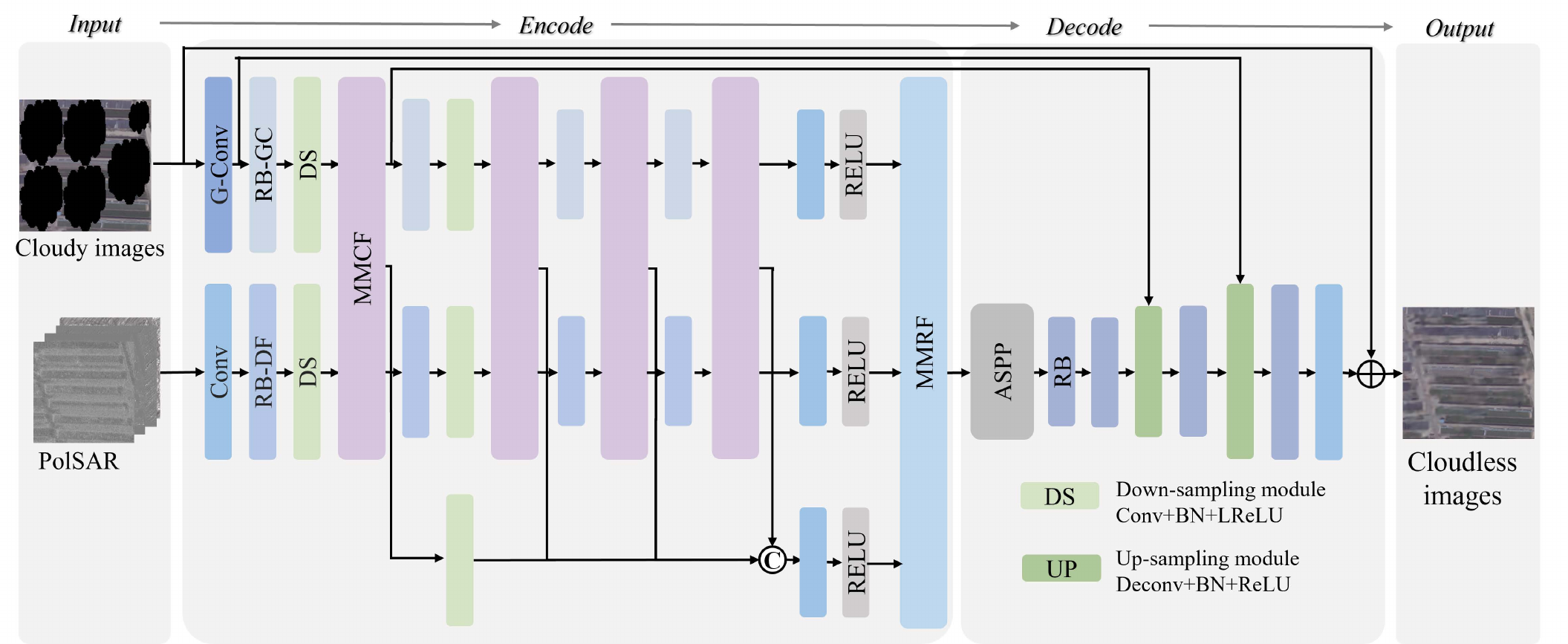}
  \caption{Overview of the proposed PolSAR-Optical data fusion based cloud removal (PODF-CR) algorithm.}
  \label{fig1}
\end{figure*}

The proposed PODF-CR algorithm utilizes simulated cloudy images and corresponding PolSAR image as inputs to reconstruct the missing regions, yielding cloud-free restoration results with appropriate spectral accuracy and high-frequency textures. As shown in Fig. \ref{fig1}, the PODF-CR model mainly consists of two parallel branches for PolSAR images and optical images. We extract polarization features images of PolSAR data based on the coherence matrix as the input of the PolSAR branch, while the cloudy optical images serve as the input for the optical branch. Additionally, cloud-free optical images are used as ground truth for loss function constraints to train the network. The overall framework of the proposed algorithm is illustrated in the following figure, mainly composed of encoding and decoding modules. 

Multi-source input images are initially encoded by the encoding layers, employing multi-level downsampling operations to expand the receptive field, effectively extracting and encoding multi-scale information to acquire more contextual details. Subsequently, within the decoding layers, the input information is fused through multi-level upsampling operations to decode the encoded features into cloud-free output images. Furthermore, prior to the final output, the input optical images are directly propagated to the output image via long skip connections to facilitate the effective transfer of cloud-free region information.

Specifically, the encoder module is a dual-branch structure that extracts PolSAR image features and optical image features, based on PolSAR image features to supplement missing optical image information. For the optical branch, the entire branch includes two convolutional layers, four optical residual blocks, two downsampling modules, four MMCF modules, a convolutional layer, and a non-linear activation layer, forming a multi-scale encoding structure. Within the optical residual blocks, the residual blocks based on gated convolutions (RB-GC) is proposed, using gated convolutions instead of ordinary convolution to capture effective information in cloud-free areas. The downsampling module mainly consists of a convolutional layer, a batch normalization layer, and an activation layer. The RB-GC and MMCF blocks maintain the input feature channels. The first convolutional layer and downsampling layer obtain images with different scales and channel numbers. At these scale levels, the number of feature channels is set to 64, 128, and 256, respectively. The downsampling module contains a 4x4 convolution with a stride of 2, aimed at reducing the size of the input image by half while doubling the number of feature channels.

Similarly, for the PolSAR image branch, the overall structure is similar to the optical branch. Considering the speckle noise in PolSAR images, the residual blocks based on dynamic filters (RB-DF) are utilized, which introduced the SCDF block in the residual blocks to suppress noise in PolSAR images. Furthermore, the fusion image obtained after the first MMCF block is downsampled to match the size of the outputs from the subsequent MMCF blocks. The output fusion features of the third branch are then created by concatenating the fusion features obtained from the MMCF block. The PolSAR images branch, optical images branch, and fusion images branch are separately passed through a convolutional layer and a non-linear activation layer, and the number of feature channels is adjusted to 256. Additionally, considering the redundancy and potential confusion of feature representations in PolSAR images, optical images, and fusion images obtained through concatenation, the features from the three branches are input into the MMRF blocks for refined features, which are then input to the decoder module.

In the decoder module, the fusion features extracted from PolSAR images and cloudy optical images are decoded to reconstruct the contaminated areas. An atrous spatial pyramid pooling (ASPP) module, two upsampling modules, four residual blocks, and a convolutional layer are contained. The upsampling module also includes a convolutional layer, a batch normalization layer, and an activation layer. Similarly, the upsampling module uses a 4x4 transposed convolution with a stride of 2 to increase the size of the input image while reducing the number of feature channels by half. The number of feature channels is set to 128 and 64 for these scale levels, respectively. The final convolutional layer ensures that the channel number of reconstructed results matches the channel number of the input image. Additionally, skip connections are performed between the downsampling and corresponding upsampling layers. This structure decreases information loss in the input data during network operation. Apart from the upsampling and downsampling modules, the kernel size of all convolutional layers is set to 3x3.

\subsection{The Encoding Module}

In the encoding module, different residual blocks, RB-GC and RB-DF, were suggested for the optical branch and the PolSAR branch, respectively. Futhermore, to better facilitate the intergration of multimodal SAR and optical images, we have introduced the MMCF module and MMRF module. With cross-modal dense connections, the MMCF block enables better interaction between PolSAR image information and optical image information, promoting feature fusion at different levels. The MMRF block, based on an attention mechanism, suppresses irrelevant information, thereby refining multimodal features and obtaining better feature representation.

\subsubsection{The Residual Blocks Based on Gated Convolutions (RB-GC)}

For the optical branch, considering the presence of a large number of invalid pixels due to clouds in optical images, in order to avoid the problems of blurring and color distortion caused by these pixels during image restoration as the network deepens, we used gated convolution to replace the ordinary convolutional layer. Gated convolution establishes a selection mechanism at different positions in the feature layer, learns the optimal mask through the network, and assigns soft encoding values to each spatial position. This enables better extraction of valid information from cloud-free areas, leading to improved feature extraction for subsequent restoration. Specifically, as shown in the Fig. \ref{fig2}(a), the RB-GC module primarily consists of three layers of 3x3 gated convolutional layers and two non-linear activation functions. Finally, it stabilizes training without introducing additional parameters through a residual scale scaling layer. In this paper, a constant value of 0.1 was chosen \citep{ref21}.

\subsubsection{The Residual Blocks Based on Dynamic Filters (RB-DF)}

\begin{figure*}[!t]
  \centering
  \includegraphics[width=\textwidth]{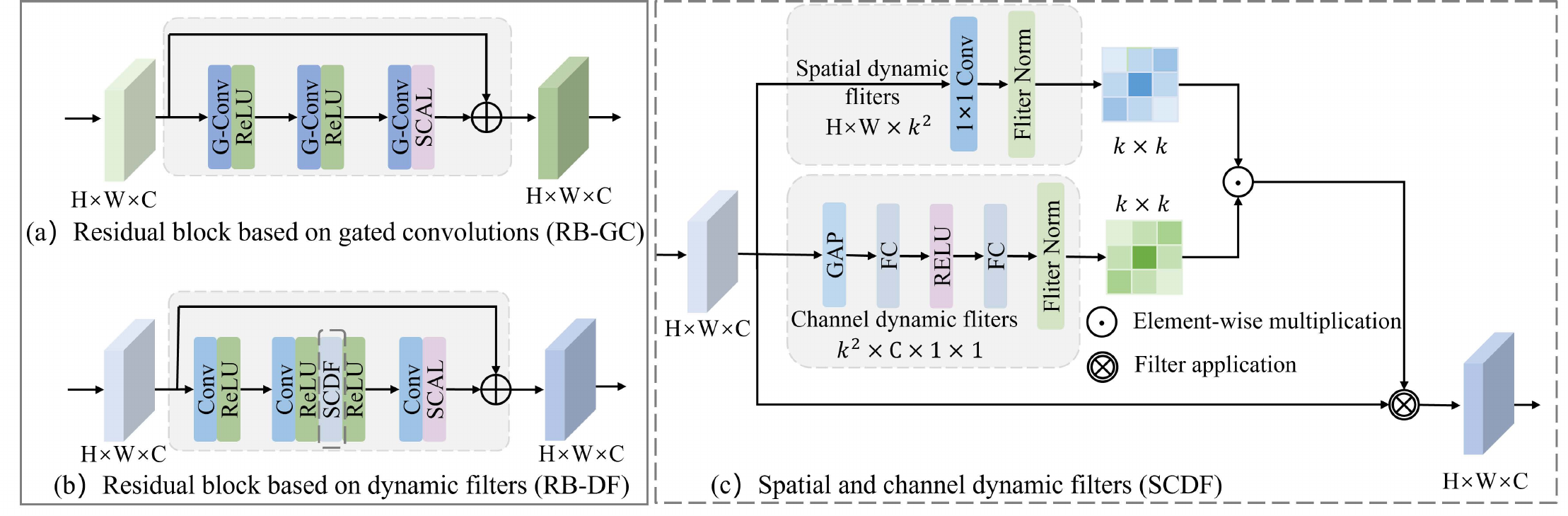}
  \caption{Illustration of the residual blocks based on gated convolutions (RB-GC), the residual blocks based on dynamic filters (RB-DF) and the spatial and channel dynamic filters (SCDF). (a) RB-GC. (b) RB-DF. (c) SCDF.}
  \label{fig2}
\end{figure*}

In the PolSAR branch, the overall structure is similar to the optical branch, based on ordinary convolutional layers for feature extraction, as shown in Fig. \ref{fig2}(b). Considering that PolSAR images are severely affected by speckle noise, we performed dynamic filtering on PolSAR features before information propagation to reduce the interference of ineffective speckle noise features in PolSAR images on the subsequent generation of optical images. Standard convolutional filters share parameters across all pixels in the image, while the values of dynamic filters are dynamically altered or predicted depending on input data. Therefore, we added the SCDF block to residual blocks which realizes dynamic filtering in both the spatial and channel dimensions, thereby aiding to suppress spatially speckle noise. The entire RB-DF module is consisting of three layers of 3×3 ordinary convolutional layers and three non-linear activation functions, with the SCDF module added after the two ordinary convolutional layers.
First, for standard convolution, given an input feature representation $F\in R^{H\times W\times C}$, with N pixels and C channels (N=H×W, where H and W are the width and height of the feature map), the standard convolution operation at a pixel can be expressed as a linear combination of input features around the pixel i:
\begin{equation}
    \label{euqa_1}
F_{(i,.)}^{'}=\sum_{j \in \Omega(i)}W F_{(j,.)}+b
\end{equation}
Where $F_{(j,.)}\in R^C$ is the feature vector at the $j^{th}$ pixel, C is the number of input feature channels, $F_{(i,.)}\in R^C$ represents the feature vector at the i-th pixel, and C is the number of output feature channels. $\Omega(i)$ is the k×k convolution window around the $i^{th}$ pixel. $W\in R^{C\times C^{'}\times k\times k}$ is the convolution filter. $b\in R^{C^{'}}$ represents the bias vector. In standard convolution, the same filter W is shared across all pixels, and the filter weights are independent of the input features.
In contrast to standard convolution, dynamic filters utilize a separate network branch to generate filters at each pixel. In this case, the spatially invariant filter W becomes a spatially variant filter $D_i\in R^{C\times C^{'}\times k\times k}$. Dynamic filters support learning of content-adaptive and flexible feature embeddings, with the parameter count being $NCC^{'}k^{2}$. In order to reduce the number of parameters while ensuring spatial and channel filtering, the SCDF module is introduced which mainly consists of spatial dynamic filters and channel dynamic filters. Specifically, the formula for the SCDF operation is as follows:
\begin{equation}
    \label{euqa_2}
F_{(i,m)}^{'}=\sum_{j \in \Omega(i)}D_i^{sp} D_m^{ch}F_{(j,m)}
\end{equation}
Where$ F_{(i,m)}^{'} \in R$ is the output feature value of the $m^{th}$ channel at the $i^{th}$ pixel. $F_{(j,m)} \in R$ is the input feature value of the $m^{th}$ channel at the j pixel. $D^{sp} \in R^{N\times k\times k}$ is the spatial filter, and $D_i^{sp} \in R^{k\times k}$ represents the filter for the $i^{th} pixel$. $D^{ch} \in R^{C\times k\times k}$ is the channel filter, and $D_m^{ch} \in R^{k\times k}$ represents the filter for the $m^{th}$ channel.

By predicting channel and spatial dynamic filters from input features, we perform the SCDF operation mentioned above to compute the output feature mapping. Comparing the general dynamic filters, it is clear that SCDF reduces the $NCC^{'}k^{2}$ sized dynamic filters into much smaller $Nk^2$ spatial and $Ck^2$ channel dynamic filters.
Fig. \ref{fig2}(c) shows the structure of the spatial and channel filter branches in the SCDF module. The spatial filter branch consists of a 1×1 convolutional layer. The dynamic channel filter branch first aggregates the input features through global average pooling, followed by a fully connected layer and a non-linear activation unit, and then generates the filter through another fully connected layer. Additionally, considering that for some input features, the generated filter values may be very large or very small, using them directly for convolution may lead to unstable training. Therefore, filter normalization (FN) is applied to the spatial and channel filters at the end.

\begin{equation}
    \label{euqa_3}
D_{i}^{sp}=\alpha^{sp}\frac{\bar{D}_{i}^{sp}-\mu(\bar{D}_{i}^{sp})}{\delta(\bar{D}_{i}^{sp})}+\beta^{sp}
\end{equation}

\begin{equation}
    \label{euqa_4}
D_{r}^{ch}=\alpha_{r}^{ch} \frac{\bar{D}_{r}^{ch}-\mu(\bar{D}_{r}^{ch})}{\delta(\bar{D}_{r}^{ch})}+\beta_r^{ch}
\end{equation}
Where $D_{i}^{sp}$ and $D_{r}^{ch}\in R^{k\times k}$ represent the spatial and channel filters generated before normalization, where $\mu(.)$ and $\delta(.)$ denote the mean and standard deviation of the filters. $\alpha^{sp}$, $\alpha_{r}^{ch}$, $\beta^{sp}$, and $\beta_{r}^{ch}$ are used for the sliding standard deviation and sliding mean and are similar to the coefficients in batch normalization (BN) \citep{ref55}. FN is able to constrain the generated filter values within a reasonable range, thus preventing the gradients from vanishing during the training process.

\subsubsection{The Multi-modality Cross Fusion (MMCF) }
\begin{figure}[!t]
\centering
\includegraphics[width=0.48\textwidth]{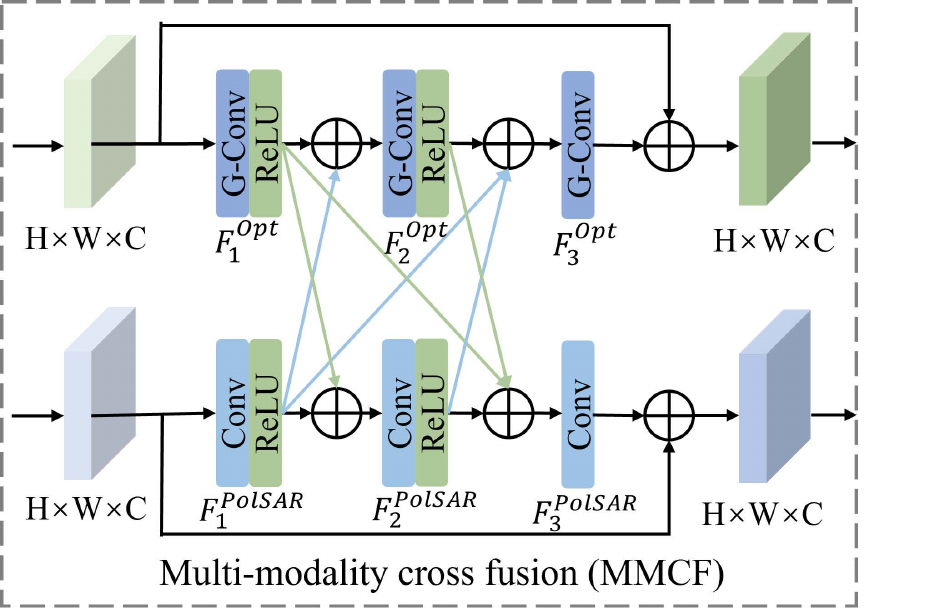}
\caption{Illustration of the multi-modality cross fusion block(MMCF).}
\label{fig3}
\end{figure}

Complementary information on PolSAR images cannot be effectively transferred across modalities, in order to better utilize cloud-free PolSAR images to guide the reconstruction of missing areas in cloudy optical images, we have introduced the MMCF block into the network. Taking into account that features at different levels contain information of different scales, receptive fields, and content, the MMCF block is applied at different stages of the encoder. This can effectively learn the complementary information between PolSAR and optical images at multiple stages, and through the interactive updating of parameters of different sub-networks, it can learn a more compact feature representation, thereby generating more comprehensive saliency features for subsequent image restoration. Unlike simple fusion with fixed periods, MMCF block performs cross-modality dense connections in the two-stream network, passing each modality's feature map to all feature maps of the other modality through skip connections, enabling cross fusion between different layers of the main network. It increases the interaction of features from different modalities, allowing fusion in both shallow and subsequent deep features, significantly improving the proportion of fusion information and better achieving interaction between PolSAR image information and optical image information, as detailed in the Fig. \ref{fig3}.

Specifically, the MMCF module has two parallel input streams, each receiving PolSAR image features and optical image features, using the basic ResNet structure, consisting of three 3x3 convolutional layers and two layers of non-linear activation functions. In the optical branch, we also use gated convolutions instead of standard convolutions for effective feature extraction. The output features after convolution are defined as $F_i^{optic}$ and $F_i^{PolSAR}$, representing the $i^{th}$ layer of optical feature block and PolSAR feature block, respectively. The detailed explanation of the MMCF fusion module is as follows, and the output after the first layer of skip fusion is as shown below:
\begin{equation}
    \label{euqa_5}
F_2^{PolSAR}=conv(F_1^{PolSAR}+(W_{12}^{Opt} F_1^{Opt}+b_{12}^{Opt}))
\end{equation}

\begin{equation}
    \label{euqa_6}
F_2^{Opt}=Gconv(F_1^{Opt}+(W_{12}^{PolSAR} F_1^{PolSAR}+b_{12}^{PolSAR}))
\end{equation}
Where $W_{ij}^{Opt}$ is the weight from the $i^{th}$ optical block to the $j^{th}$ PolSAR feature block, and $b_{ij}^{Opt}$ is the corresponding bias. $W_{ij}^{PolSAR}$ is the weight from the $j^{th}$ optical block to the $i^{th}$ PolSAR feature block. 

Similar to the first layer skip fusion, the output for the second layer skip fusion is as follows:
\begin{equation}
    \label{euqa_7}
\begin{split}
F_3^{PolSAR}=conv(F_2^{PolSAR}+(W_{13}^{Opt} F_1^{Opt}+b_{13}^{Opt} )\\
+(W_{23}^{Optic} F_2^{Opt}+b_{23}^{Opt}))
\end{split}
\end{equation}

\begin{equation}
    \label{euqa_8}
\begin{split}
F_3^{Optic}=Gconv(F_2^{Opt}+(W_{13}^{PolSAR} F_1^{PolSAR}+b_{13}^{PolSAR} )\\
+(W_{23}^{PolSAR} F_2^{PolSAR}+b_{23}^{PolSAR} ))
\end{split}
\end{equation}

\subsubsection{The Multi-modality Refinement Fusion (MMRF)}
In order to better extract the fused features of PolSAR images and optical images, we have introduced a MMRF module in terms of feature selection and enhancement. Through attention mechanisms, unnecessary information is suppressed, and complementary information is highlighted to guide the fusion of PolSAR images and optical images, refining the extracted multi-modal features. Considering that each modality features extracted contains rich information, common fusion strategies such as concatenation may introduce information redundancy, potentially confusing feature representations and increasing the difficulty of learning effective feature representations. Therefore, we have introduced spatial-channel attention units (SCRU) to suppress irrelevant information. Additionally, to better utilize the complementarity and correlation between PolSAR images and optical images, we have introduced multi-modality weighted refinement units (MWRU) to capture long-term dependencies between PolSAR images and optical images and refine the features of fusion images from a global perspective, enhancing the consistency of the reconstructed images.

\subsubsubsection{The Spatial-channel Refinement Unit (SCRU)}
Firstly, PolSAR data, optical data, and fusion data all contain rich spatial and channel information, which may lead to information redundancy. This can pose problems for effective feature representation in image restoration. The attention module can better utilize global feature information, learning the relationships between specific positions and all other positions, and is widely used in the field of image restoration \citep{ref22, ref23}. Therefore, we refine the features obtained from each image in terms of spatial and channel dimensions using the SCRU to obtain effective feature representations.
 
\begin{figure*}[!t]
  \centering
  \includegraphics[width=\textwidth]{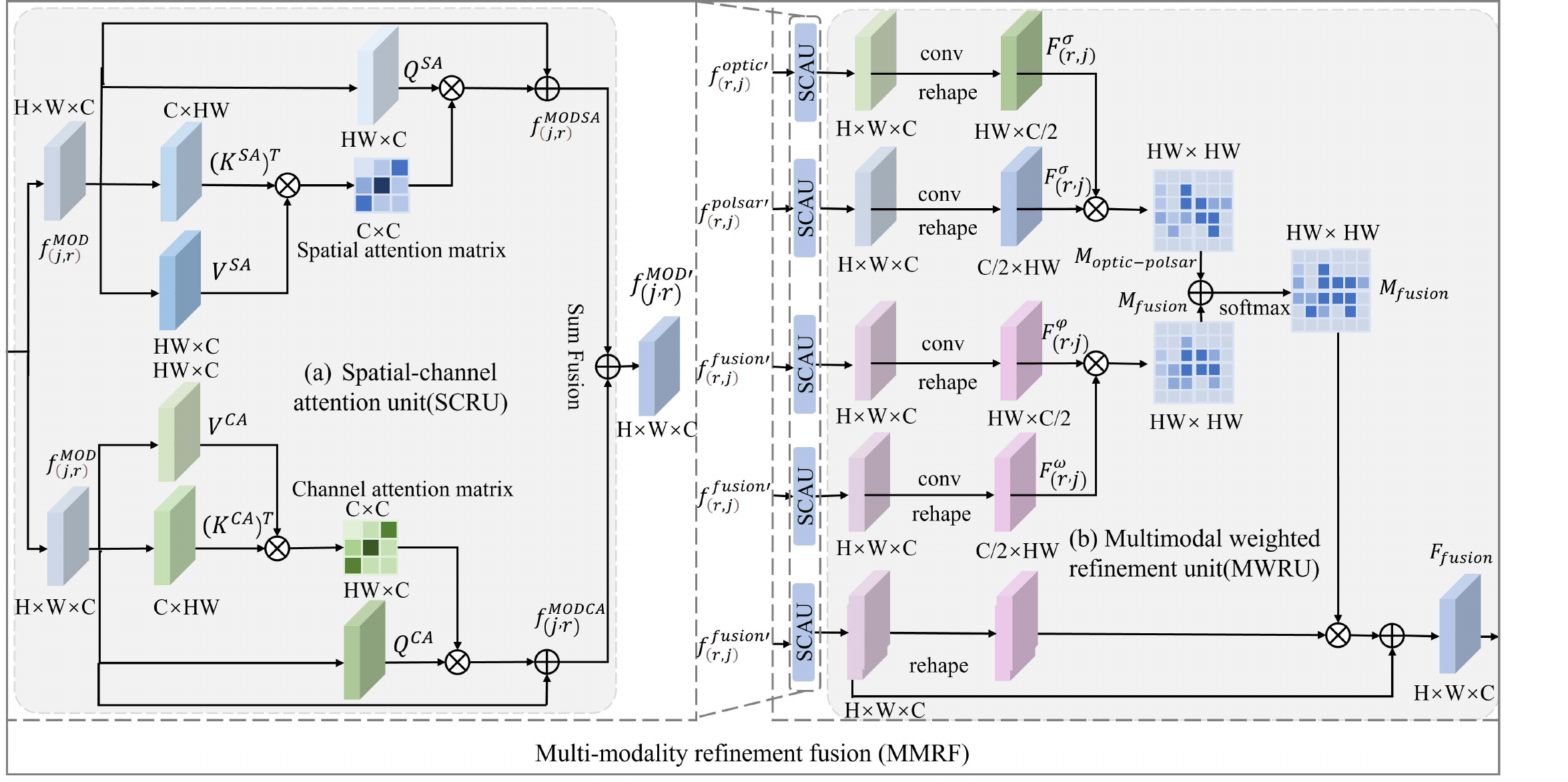}
  \caption{Illustration of the multi-modality refinement fusion. (a) Multi-modal refinement fusion block (MRFB). (b) Spatial-channel attention unit(SCAU).}
  \label{fig4}
\end{figure*}

Following the encoding module, features from the PolSAR pictures, the optical images, and the fusion images are produced by performing basic concatenation operations are obtained. We integrate spatial and channel attention mechanisms to process the data from these three branches and suppress redundant information. The features obtained through channel attention and spatial attention are fused to generate the refined features. The spatial attention module is replaced by an efficient spatial attention mechanism to reduce computational costs. As shown in the Fig. \ref{fig4}(a), the data of the three encoder branches are input into SCRU separately, and the corresponding spatial and channel attention maps are calculated. Then, they are merged with the attention maps through matrix multiplication, and the spatial and channel refined features are obtained through residual connections. Finally, the spatial and channel features are fused to generate the final output. The calculation process for the spatial attention $f_{(j,m)}^{MODSA}$ is as follows:
\begin{equation}
    \label{euqa_9}
f_{(j,m)}^{MODSA}=Q^{SA} (K^{SA})^T V^{SA}+f_{(j,m)}^{MOD}
\end{equation}

Where $f_{(j,m)}^{MOD}$ represents the input feature of the $j^{th}$ pixel in the $m^th$ channel $K^{SA}$, $Q^{SA}$, $V^{SA}$ are obtained by a convolution layer and a reshaping operation of input features, and $MOD\in{optic,PolSAR,fusion}$ refers to the features of the optical images, the PolSAR images, and the fused images, respectively.
Like spatial attention, $K^{CA}$, $Q^{CA}$, $V^{CA}$ are obtained by a reshaping operation of input features $f_{(j,m)}^{MOD'}$, the channel attention $f_{(j,m)}^{MODCA}$ calculation process is as follows:
\begin{equation}
    \label{euqa_10}
f_{(j,m)}^{MODCA}=Q^{CA} (K^{CA})^T V^{CA}+f_{(j,m)}^{MOD}
\end{equation}

Finally, the refined features $f_{(j,m)}^{MOD'}$ are obtained by fusing the features obtained through spatial attention and channel attention.
\begin{equation}
    \label{euqa_11}
f_{(j,m)}^{MOD'}=f_{(j,m)}^{MODSA}+f_{(j,m)}^{MODCA}
\end{equation}

\subsubsubsection{The Multi-modality Weighted Refinement Unit (MWRU)}
Considering the complementarity and correlation between optical and PolSAR images, the features of optical images, fusion images, and PolSAR images through the MWRU to capture long-range dependencies and supplementary details information between different modal data, thus obtaining globally refined fusion features, and achieving better fusion of the PolSAR images and the optical images. The specific structure of the MWRU is shown in Fig. \ref{fig4}(b) below. The features from each branch, $f_{(j,m)}^{MOD'}\in R^{H\times W\times C}$, which are output by SCAU, are embedded into the MWRU. Firstly, we use a convolutional layer to reduce the number of channels and map different modalities to a unified feature space, which can be represented as:
\begin{equation}
    \label{euqa_12}
F_{(j,m)}^\theta=W_\theta f_{(j,m)}^{optic'}
\end{equation}
\begin{equation}
    \label{euqa_13}
F_{(j,m)}^\sigma=W_\sigma f_{(j,m)}^{polsar'}
\end{equation}
\begin{equation}
    \label{euqa_14}
F_{(j,m)}^\varphi=W_\varphi f_{(j,m)}^{fusion'}
\end{equation}
\begin{equation}
    \label{euqa_15}
F_{(j,m)}^\omega=W_\omega f_{(j,m)}^{fusion'}
\end{equation}
Where $W_\theta$ and $W_\sigma$ represent the learnable embedding weights of the convolutional layers on the optical branch and PolSAR branch, and $W_\varphi$, and $W_\omega$ represent represent the learnable embedding weights of the convolutional layers on the fusion branch. $F_{(j,m)}^\theta$ and $F_{(j,m)}^\sigma$ respectively represent the output features through SCAU on the optical branch and the PolSAR branch, and $f_{(j,m)}^{fusion'}$ represents the output features through SCAU on the fusion branch.

Then, the correlation between the optical images features $F_{(j,m)}^\theta$ and the PolSAR images features $F_{(j,m)}^\sigma$, as well as the self-correlation of the fusion images features $F_{(j,m)}^\varphi$ and $F_{(j,m)}^\omega$, are computed in a pixel-wise manner.
\begin{equation}
    \label{euqa_16}
M_{optic-polsar}=softmax(F_{(j,m)}^\theta \otimes F_{(j,m)}^\sigma)
\end{equation}

\begin{equation}
    \label{euqa_17}
M_{fusion}=softmax(F_{(j,m)}^\varphi \otimes F_{(j,m)}^\omega)
\end{equation}
Where $\otimes$ denotes matrix multiplication, and softmax is the activation function. $M_{optic-polsar}\in R^{HW\times HW}$ emphasizes the correlation between optical images features and PolSAR images features, and $M_{fusion}\in R^{HW\times HW}$ simulates the dependency relationship within the fusion images features.

Finally, these two correlation information jointly generate global dependency weights to refine the original input features:
\begin{equation}
    \label{euqa_18}
\begin{split}
F_{fusion}=R(f_{(j,m)}^{fusion'})\otimes softmax(M_{optic-polsar}\otimes M_{fusion})\\
+f_{(j,m)}^{fusion'}
\end{split}
\end{equation}
Where R denotes the reshape operation, reshaping the features from $R\in {C\times H\times W} to R\in {C\times HW}$. The cross-modal global dependency weights generated by $M_{optic-polsar}\otimes M_{fusion}$ refine the original input modal features from a global perspective, allowing for a better representation of fusion features and thus enhancing the restoration accuracy.

\subsubsection{The Decoding Module}
\begin{figure}[!t]
\centering
\includegraphics[width=0.48\textwidth]{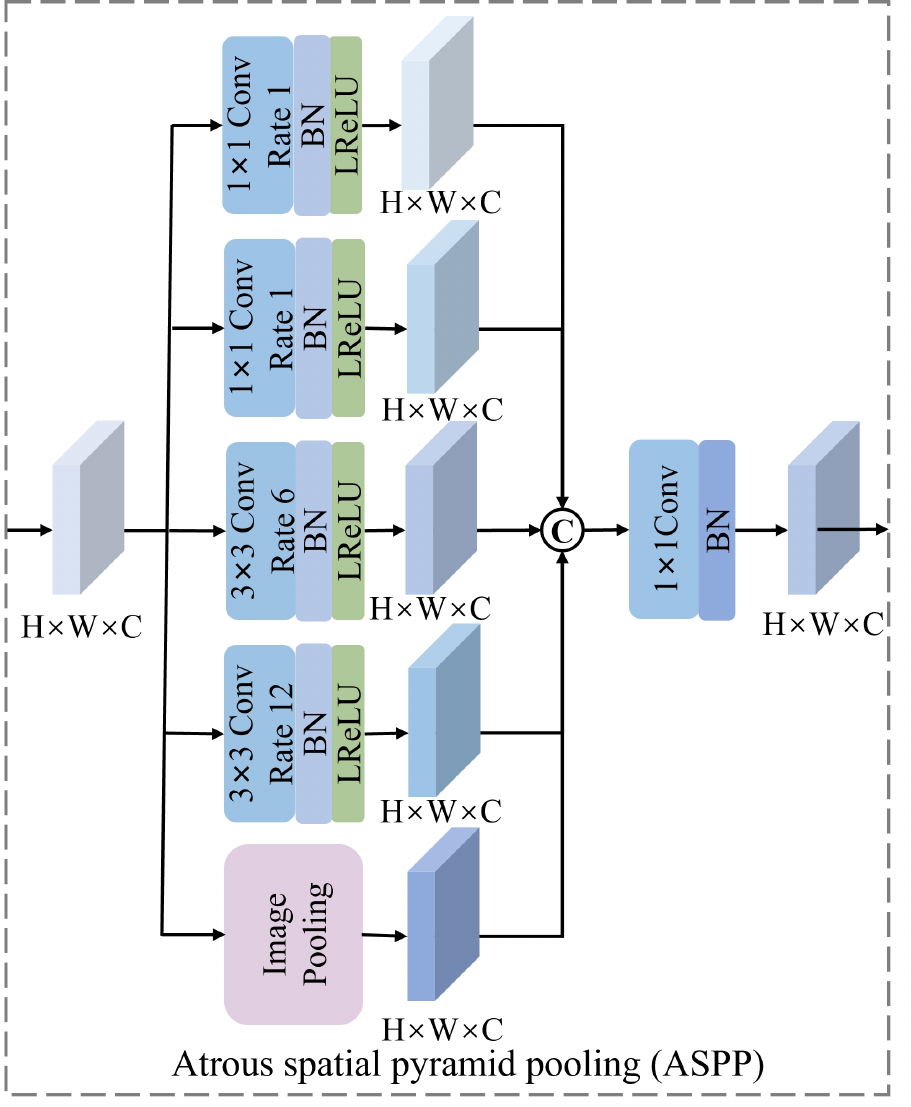}
\caption{Illustration of the atrous spatial pyramid pooling (ASPP) structure.}
\label{fig5}
\end{figure}

Remote sensing images contain objects of different scales, and the reconstruction of missing images relies on information from different scales. We implemented the ASPP structure to improve the utilization of spatial multi-scale information. The ASPP structure, originating from the DeepLabv3+ paper \citep{ref56}, typically consists of multiple parallel convolution branches, each using dilated convolutions with different dilation rates to capture context information at different scales. This has been widely used in the field of image restoration \citep{ref57}. As shown in Fig. \ref{fig5}, the ASPP module is based on a 1×1 convolution and three 3×3 atrous convolutions to extract multi-scale feature information, while also employing global pooling to preserve global context information. The dilation rates of the atrous convolutions are 6, 12, and 18, respectively. Subsequently, the extracted features are concatenated and passed through a 1×1 convolutional layer prior to the residual block. Extracting multi-scale features from the fused image after encoding and improving the circumstances for the restoration of cloud-free areas.

\subsubsection{Loss Function}
In this model, in order to achieve good results, we use the L1 metric (mean absolute error) as the fundamental error function to globally constrain the reconstruction. Let the predicted output image be denoted as $I_P$ and the cloud-free target image as $I_T$. Based on the reconstruction loss between the prediction and the target on the global image, the per-pixel reconstruction loss $L_g$ can be expressed as:
\begin{equation}
    \label{euqa_19}
L_g=1/N \parallel I_T-I_P\parallel_1
\end{equation}
Where N is the total number of pixels across all bands in the optical image.

Additionally, a loss function should be designed to focus on the reconstruction of missing parts. Therefore, we construct a local loss function $L_l$ using the cloud mask M:
\begin{equation}
    \label{euqa_20}
L_l=1/N \parallel MI_T-MI_P\parallel_1
\end{equation}
Furthermore, considering human visual perception, we choose the structural similarity (SSIM) loss function to measure the structural similarity between the two images, preserving more high-frequency information.
\begin{equation}
    \label{euqa_21}
L_{ssim}=1-SSIM(I_T,I_P)
\end{equation}

\begin{equation}
\begin{aligned}
    \label{euqa_22}
SSIM&(I_T,I_P)=\\
&\frac{(2\mu_{I_{T}} \mu_{I_{P}}+C_1)(2\sigma_{I_{T} I_{P}}+C_2)}{({\mu_{I_{T}}}^2+{\mu_{I_{P}}}^2+C_1)({\sigma_{I_{T}}}^2+{\sigma_{I_{P}}}^2+C_2)}.
\end{aligned}
\end{equation}
Where $\mu_{I_{T}}$ and $\mu_{I_{P}}$ are the mean values of the restored optical image and the ground truth image, $\sigma_{I_{T}}$ and $\sigma_{I_{P}}$ are the standard deviations of the two images, and $\sigma_{I_{T} I_{P}}$ is the covariance of the two images. $C_1$ and $C_2$ are constants.
Finally, the total loss function for the cloud removal network is obtained as:
\begin{equation}
    \label{euqa_23}
L_{cr}=L_g+\lambda_1 L_l+\lambda_2 L_{ssim}
\end{equation}
Where $\lambda_1$ and $\lambda_2$ are the weights for $L_l$ and $L_{ssim}$, respectively.

\section{Experiments}\label{Experiments}

To verify the feasibility of our network on the cloud removal task, we collected an PolSAR based multi-source dataset for experiments. Details of our experiments are given as below.

\subsection{Experiment Settings}
\subsubsection{OPT- BCFSAR -PFSAR Dataset}

The optical images and SAR images in the OPT-BCFSAR-PFSAR dataset are both obtained from the Zhangye area in Gansu Province, captured on September 12, 2021. The airborne optical data includes RGB and near-infrared channels with a spatial resolution of 0.2 meters. The airborne C-band full-polarization single-look complex SAR data includes HH, HV, VH, and VV polarizations. To ensure pixel-wise correspondence between the optical and PolSAR images, the optical images were resampled to a ground resolution of 0.5 meters to match the PolSAR images. Since the HV polarization band is consistent with the VH polarization band, only the HV polarization band is used. The single-look complex PolSAR data was transformed into the backscatter coefficient feature images (BCFSAR) for the three bands, and corresponding the polarization feature images (PFSAR) were extracted for 9 bands.

The optical data and PolSAR data were registered with a deviation of less than 2 pixels. Subsequently, the registered SAR and optical images were cropped into small regions of size 256px × 256px, with a step size of 128px. The cropped dataset was manually annotated for categories, and optical images with varying cloud cover were simulated using randomly generated cloud masks, which led to the first remote sensing image dataset based on polarization features. The OPT-BCFSAR-PFSAR dataset comprises corresponding image pairs, each consisting of the cloudy optical image, the cloud-free optical image, the BCFSAR image, and the corresponding PFSAR image from the same area, totaling 2030 image pairs. The information is detailed in Table \ref{tab:table_1}.

\begin{table*}[htbp]
  \centering
  \caption{Description of OPT-BCFSAR-PFSAR dataset.}
  \label{tab:table_1}
  \renewcommand{\arraystretch}{1.3}
  \begin{tabular}{cccc}
  \toprule
  Data & Spatial resolution(m) & Channel number & Patch size(px)\\
  \midrule
Cloudy optical images & 0.5 & 4 & 256 \\
Cloudless optical images & 0.5 & 4 & 256 \\
BCFSAR images & 0.5 & 3 & 256 \\
PFSAR images & 0.5 & 9 & 256 \\
  \bottomrule
  \end{tabular}
  \end{table*}

The dataset has been manually annotated and is categorized and the classification system was defined as the following 7 land-cover classes: “Barren and sparse vegetation”, “Building”, “Cropland”, “Tree cover”, “Grassland”, “Traffic route”, and “Water”. 

\begin{figure*}[!t]
  \centering
  \includegraphics[width=\textwidth]{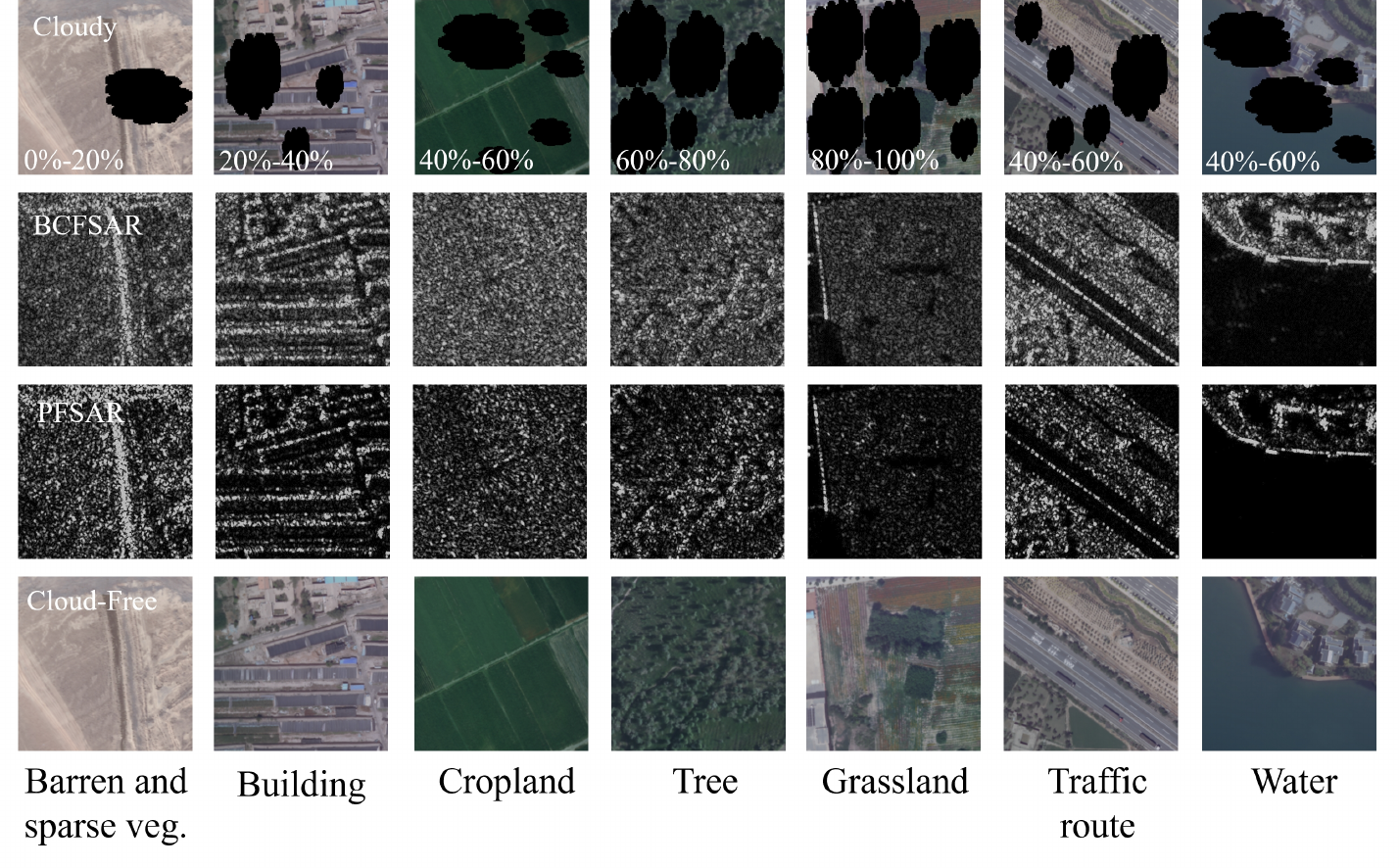}
  \caption{Example of the main categories of the OPT-BCFSAR-PFSAR dataset.}
  \label{fig6}
\end{figure*}

The cloud coverage percentage of the dataset was divided into 0-20\%, 20-40\%, 40-60\%, 60-80\% and 80-100\%. In order to evaluate the performance of image restoration, the images of each cloud cover were randomly assigned in a ratio of 1:1:1:1 to generate simulated cloud maps within each category. Examples of images from the dataset are displayed in the Fig. \ref{fig6}. 

The dataset consists of a total of 2030 image pairs, with 80\% of the images used for the training dataset and the remaining 20\% used for testing the model. Therefore, the training set consists of 1630 image pairs, and the test set consists of 400 image pairs. The total number of samples for each class and the corresponding division into training and testing samples are shown in the Table \ref{tab:table_2}. To ensure the stability of model training, optical data and PolSAR data are normalized before being input into the model to ensure that the data is within the same range.

\begin{table*}[htbp]
  \centering
  \caption{Numbers of different samples in OPT-BCFSAR-PFSAR dataset.}
  \label{tab:table_2}
  \renewcommand{\arraystretch}{1.3}
  \begin{tabular}{cccc}
  \toprule
  Land cover & Samples & Training samples & Testing samples\\
  \midrule
Barren and sparse vegtation & 310 & 250 & 60 \\
Building & 395 & 315 & 80 \\
Cropland & 380 & 305 & 75 \\
Tree cover & 230 & 185 & 45 \\
Grassland & 85 & 70 & 15 \\
Traffic route & 245 & 195 & 50 \\
Water & 385 & 310 & 75 \\
  \bottomrule
  \end{tabular}
  \end{table*}

\subsubsection{Evaluation Metrics}

The cloud removal performance of the normalized data is evaluated based on peak signal-to-noise ratio (PSNR), structural similarity index (SSIM), correlation coefficient (CC), and spectral angle mapper (SAM). The pixel-level reconstruction performance of the images is evaluated using PSNR metrics and SSIM reflects the spatial structural restoration based on visual perception principles. CC is a measure of correlation between images, while SAM retains spectral information in the reconstructed results. Higher PSNR, CC, and SSIM values, as well as lower SAM values, indicate better image reconstruction quality.

\subsubsection{Implementation Details}

The cloud removal network proposed in this paper is optimized using the Adaptive Moment Estimation (ADAM) algorithm \citep{ref58}, with the hyperparameters set to $\beta1$=0.5 and $\beta2$=0.999. After systematic search, the optimal learning rate was found to be $7\times10^{-5}$. For our model and other deep learning-based methods, the training process involves 200 iterations. The empirical weights are set to $\lambda1=10$ and $\lambda2=1$. The algorithm is implemented using the PyTorch framework on a Windows 10 environment, utilizing an NVIDIA GeForce RTX 3090 GPU.

\subsubsection{Compared Algorithms}

We have selected six deep learning-based algorithms to compare with our method: the SpAGAN model \citep{ref11}, the Pix2pix model \citep{ref16}, the SAR-Opt-cGAN model \citep{ref19}, the DSen2-CR model \citep{ref21}, the USSRN-CR model \citep{ref23}, and the GLF-CR model \citep{ref49}. These models are used to remove clouds from optical images using PolSAR images. The SpAGAN model is based on attention mechanisms to search for similar patches in cloud-free areas of contaminated images for restoration. The Pix2pix model uses a U-Net architecture to directly convert PolSAR images into optical images. The SAR-opt-cGAN model uses a cGAN network to generate cloud-free optical data from cloud-damaged optical input and auxiliary SAR images. The DSen2-CR model and the USSRN-CR model take concatenated PolSAR data and degraded optical images as input and use residual networks to obtain cloud-free optical images. The GLF-CR model combines cross-fusion with spatial self-attention mechanisms on top of the residual network, using complementary information in PolSAR images to guide global and local information reconstruction. The aforementioned algorithms and the method proposed in this paper are trained on the same dataset. 

\subsection{Comparison with the State of the Art}
\subsubsection{Analysis on the OPT-BCFSAR-PFSAR Dataset}

\begin{figure*}[!t]
  \centering
  \includegraphics[width=\textwidth]{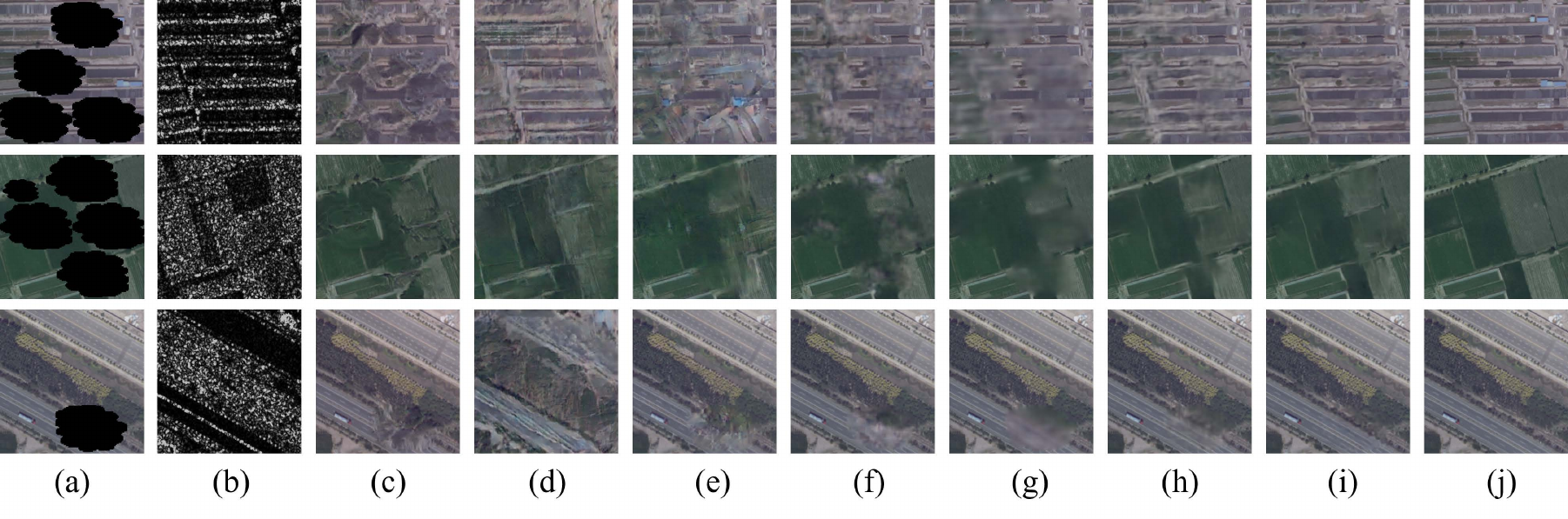}
  \caption{Experimental results of the PODF-CR model to state-of-the-art methods in dataset. (a) Cloudy images. (b) PFSAR images. (c) Results of the SpAGAN model. (d) Results of the pix2pix model. (e) Results of the SAR-opt-cGAN model. (f) Results of the DSen2-CR model. (g) Results of the GLF-CR model. (h) Results of the USSRN-CR model. (i) Results of the PODF-CR model. (j) Ground truth.}
  \label{fig7}
\end{figure*}

To verify the advantage of the PODF-CR algorithm in utilizing PolSAR image capabilities, we compared it with six other advanced algorithms on the OPT-BCFSAR-PFSAR dataset. The experimental results are shown in Fig. \ref{fig7}. The simulated damaged image is shown in Fig. \ref{fig7}(a) and the PFSAR image in Fig. \ref{fig7}(b). The results of the SpAGAN model, Pix2pix model, SAR-opt-cGAN model, DSen2-CR model, GLF-CR model, USSRN-CR model, and the proposed model are shown in Fig. \ref{fig7}(c)-(i). Fig. \ref{fig7}(j) displays the ground truth. Additionally, quantitative results are presented in Table \ref{tab:table_3}, including PSNR, SSIM, CC, and SAM metrics. Compared to state-of-the-art methods, the proposed PODF-CR network brings significant improvements.

Based on the different auxiliary information used, the comparative algorithms can be categorized as the SpAGAN model, which relies solely on missing optical data, the Pix2pix model, which transforms PolSAR images into cloud-free optical images, and other models that utilize both PolSAR images and missing optical images. Firstly, as PolSAR images encode rich geometric information in cloud-covered areas, they provide relevant information for the reconstruction of land features. The SpAGAN model, which relies solely on multi-cloud optical images, performs poorly compared to models incorporating PolSAR data. Therefore, in the first and second rows of Fig. \ref{fig7}, for large missing areas such as buildings or cropland, the SpAGAN model generates pseudo-shadows in cloud areas, resulting in low similarity to the cloud-free areas. The Pix2pix model, relying solely on PolSAR images for reconstruction, can highlight the geometric features of land classes. However, due to the lack of information from cloud-free areas in the simulated optical images, significant changes occur in the cloud-free areas, as shown in the third row, where the reconstructed traffic route in the cloud-free area deviates from the ground truth image. Additionally, the reconstructed results from Pix2pix also exhibit significant spectral deviations, with the SAM metric reaching its highest value.

The SAR-opt-cGAN model and DSen2-CR model simultaneously use simulated images and PolSAR images as input. Compared to the previous two models, the joint use of both inputs improves the accuracy of the reconstructed images and also shows better performance in terms of spectral characteristics. However, they still struggle to accurately reconstruct the texture details of land features in the face of large missing areas, resulting in a low similarity to the structure of real images. The GLF-CR model and USSRN-CR model introduce attention mechanisms to capture global information, thereby maintaining better global consistency in the reconstructed images. However, the reconstructed areas still exhibit blurry image textures. In contrast, our model excels in the reconstruction of structure information and texture details, achieving better restoration results and preserving the overall color tone. For example, in the first row of Fig. \ref{fig7}, both the USSRN-CR model and the GLF-CR model fail to accurately reconstruct the clear structure of buildings, while our model can restore the overall outline of buildings, capturing more high-frequency details, and improving the reconstruction accuracy.

\begin{table}[htbp]
  \centering
  \caption{Quantitative comparisons of proposed PODF-CR methods to state-of-the-art methods. The best performance is shown in \textbf{bold}}
  \label{tab:table_3}
  \renewcommand{\arraystretch}{1.3}
  \begin{tabular}{ccccc}
  \toprule
  Method & PSNR$\uparrow$ & SSIM$\uparrow$ & CC$\downarrow$ & SAM$\downarrow$\\
  \midrule
SpAGAN & 27.347	&0.767	&0.759	&2.916 \\
Pix2pix	&23.166	&0.506	&0.505	&4.955\\
SAR-Opt-cGAN	&28.000	&0.773	&0.792	&2.383\\
DSen2-CR	&31.005	&0.843	&0.840	&1.248\\
GLF-CR	&32.383 	&0.856 	&0.873 	&1.231 \\
USSRN-CR	&33.036	&0.867	&0.885 	&1.120\\
PODF-CR	&\textbf{34.992}	&\textbf{0.892}	&\textbf{0.922}	&\textbf{0.916}\\
  \bottomrule
  \end{tabular}
  \end{table}

\subsubsection{Analysis on Different Land Cover and Cloud Cover Levels}

\begin{figure*}[!t]
  \centering
  \includegraphics[width=\textwidth]{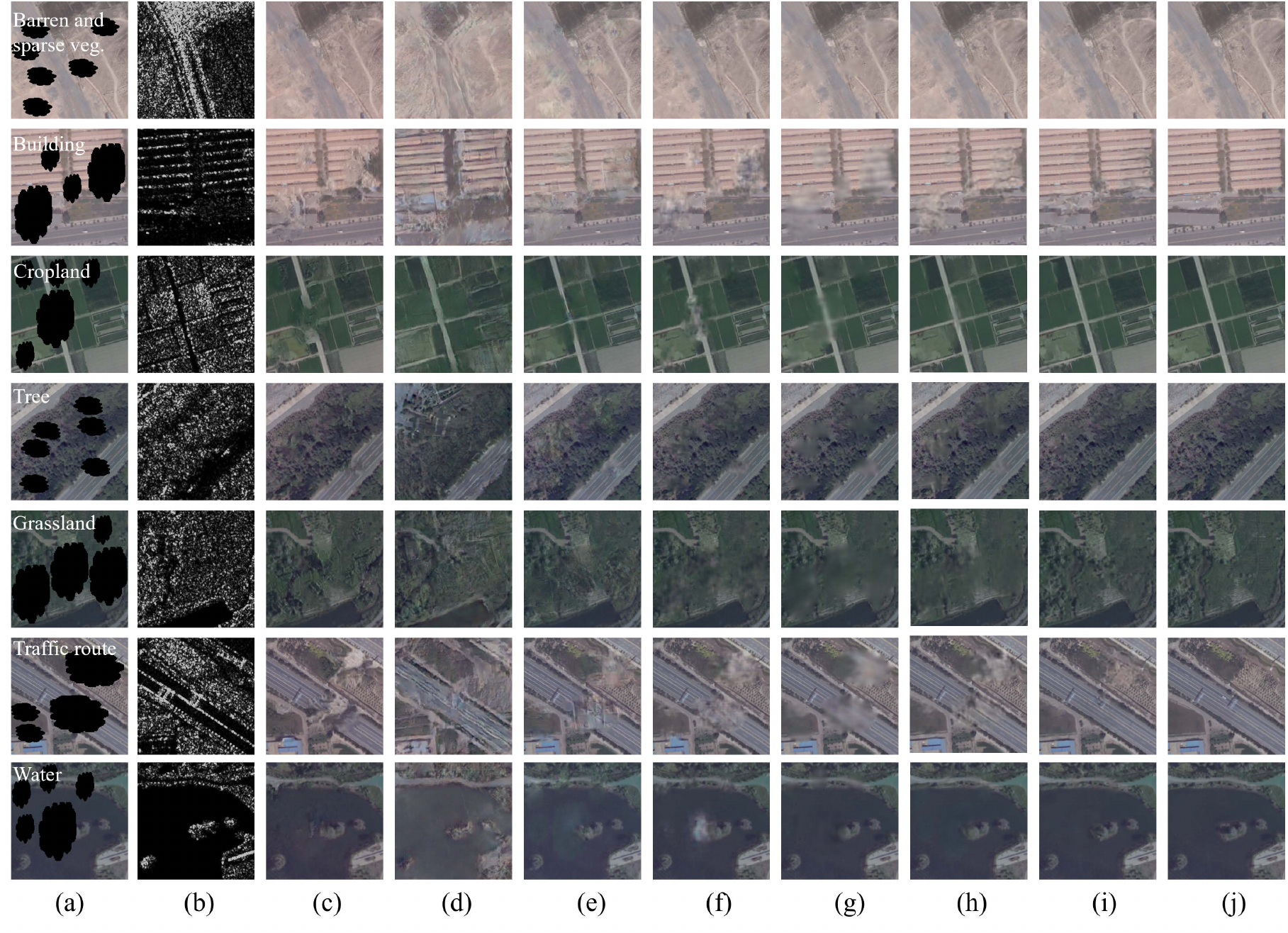}
  \caption{Experimental results of the PODF-CR model to state-of-the-art methods based on different category in dataset. (a) Cloudy images. (b) PFSAR images. (c) Results of the SpAGAN model. (d) Results of the pix2pix model. (e) Results of the SAR-opt-cGAN model. (f) Results of the DSen2-CR model. (g) Results of the GLF-CR model. (h) Results of the USSRN-CR model. (i) Results of the PODF-CR model. (j) Ground truth.}
  \label{fig8}
\end{figure*}

We compared the PODF-CR network with other state-of-the-art cloud removal methods under different land cover categories, including barren and sparse vegetation, building, cropland, tree cover, grassland, traffic route, and water, to evaluate the reconstruction effectiveness. Fig. \ref{fig8} shows the reconstruction results of various cloud removal algorithms based on each category, with the cloud coverage ranging from 40\% to 60\%. Fig.\ref{fig8}(a)-(b) depict the cloudy images and PFSAR images. The results of the SpAGAN model, Pix2pix model, SAR-opt-GAN model, DSen2-CR model, GLF-CR model, USSRN-CR model, and the proposed model are shown in Fig. \ref{fig8}(c)-(i). Fig. \ref{fig8}(j) represents the ground truth. 

Overall, in terms of texture details, all algorithms achieve better reconstruction results on large and uniform land cover categories, such as the water category. However, for complex textured features such as building, the visual effects are relatively poor. As shown in the third row of Fig. \ref{fig8}, it can be seen that the reconstruction results of the SpAGAN model and the Pix2pix model changed greatly, while the SAR-opt-cGAN model, the DSen2-CR model, and the GLF-CR model produce blurry reconstruction results. Only the USSRN-CR model and the PODF-CR model accurately reconstruct the geometric structure of cropland, and the generated restored images have more texture details. In Fig. \ref{fig8}, it is evident that the proposed method produces better qualitative results for the construction of continuous linear features such as building and traffic route, the reconstruction of texture details of ground objects such as tree cover and grassland, and the restoration of uniform features such as barren and sparse vegetation, cropland, and water.

\begin{figure*}[!t]
  \centering
  \includegraphics[width=\textwidth]{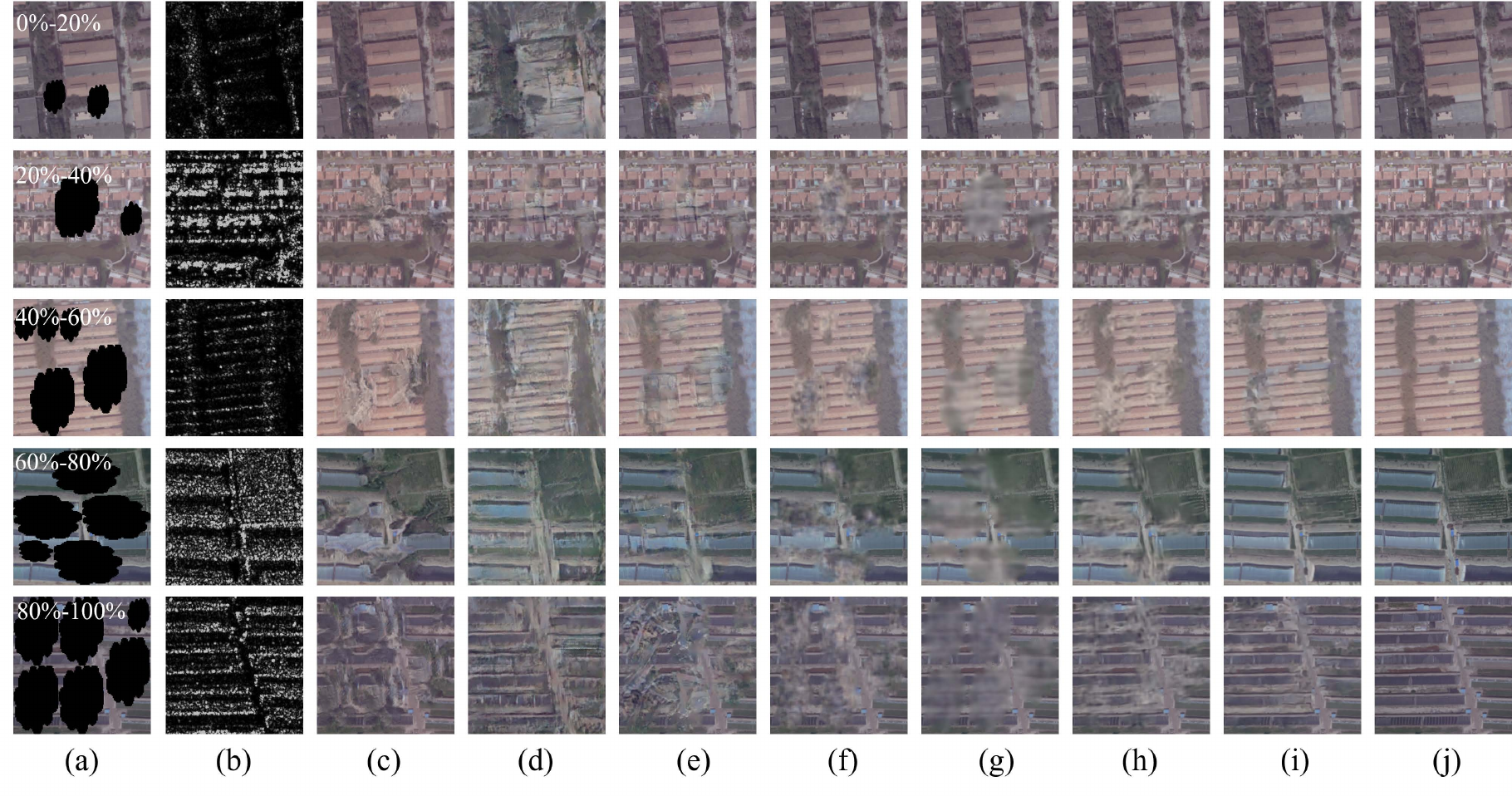}
  \caption{Experimental results of the PODF-CR model to state-of-the-art methods based on different cloud coverage in dataset. (a) Cloudy images. (b) PFSAR images. (c) Results of the SpAGAN model. (d) Results of the pix2pix model. (e) Results of the SAR-opt-cGAN model. (f) Results of the DSen2-CR model. (g) Results of the GLF-CR model. (h) Results of the USSRN-CR model. (i) Results of the PODF-CR model. (j) Ground truth.}
  \label{fig9}
\end{figure*}

\begin{figure*}[!t]
  \centering
  \includegraphics[width=\textwidth]{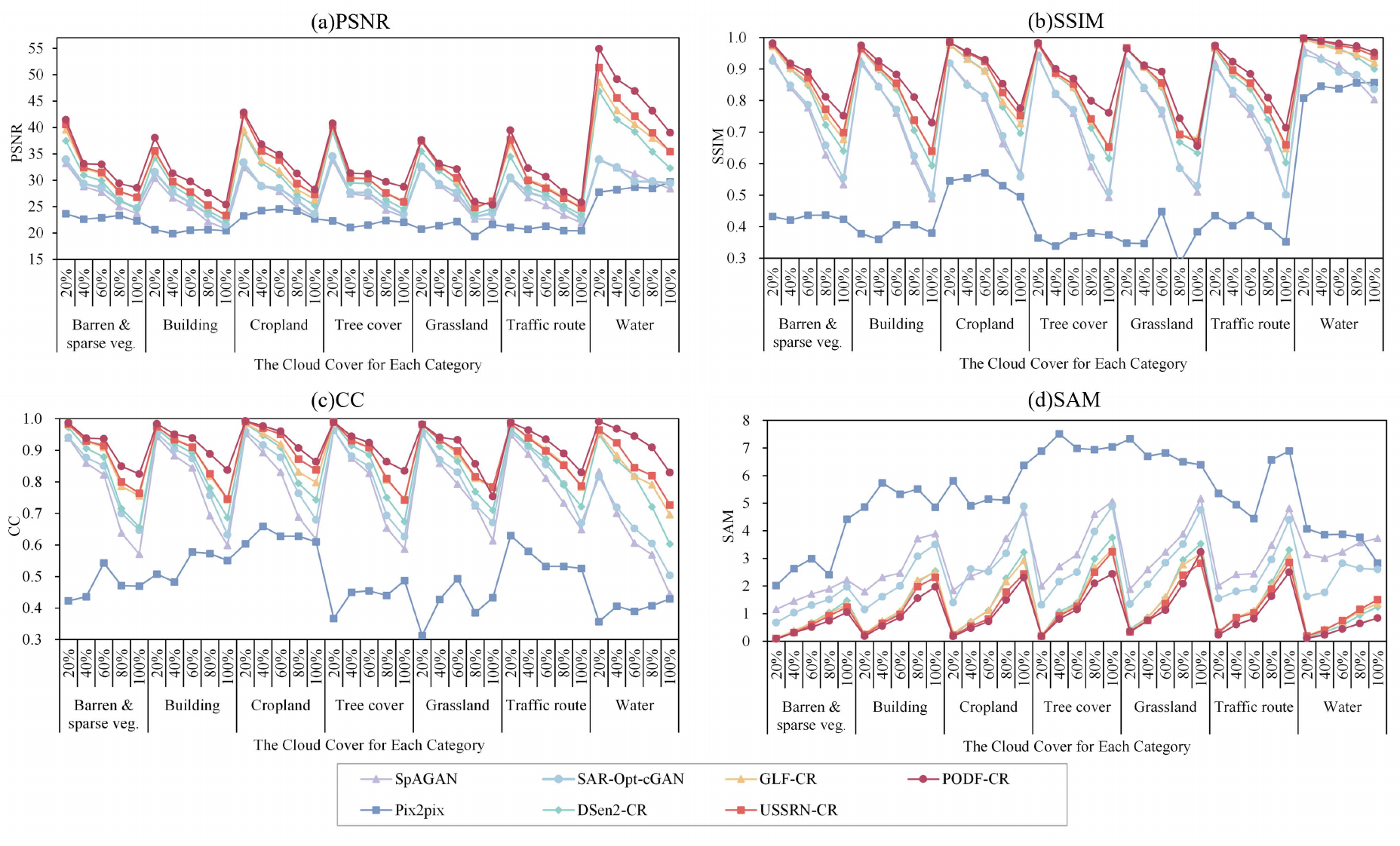}
  \caption{Comparisons of the PODF-CR model to state-of-the-art methods on different category and cloud coverage in terms of the PSNR, SSIM, SAM, and MAE quality metrics.}
  \label{fig10}
\end{figure*}

We further compared the proposed model with other cloud removal methods at different cloud cover levels. We evaluated the performance of cloud removal on images with cloud cover ranging from 0\% to 20\%, 20\% to 40\%, 40\% to 60\%, 60\% to 80\%, and 80\% to 100\%. Fig. \ref{fig9} shows the reconstruction results for the building category based on various cloud cover levels. Fig. \ref{fig9}(a) displays the damaged optical images. Fig. \ref{fig9}(b) displays the PFSAR images. Fig. \ref{fig9}(c)-(i) show the reconstruction results for the SpAGAN model, Pix2pix model, SAR-opt-cGAN model, DSen2-CR model, GLF-CR model, USSRN-CR model, and the proposed model. Fig. \ref{fig9}(j) represents the ground truth. Fig. \ref{fig10} presents the comparison results of quality metrics including PSNR, SSIM, CC, and SAM at different categories and cloud cover levels.

The performance of all models is negatively correlated with cloud cover, except for the Pix2pix model, which solely relies on PolSAR data input to generate cloud-free images. All models achieve the best visual effects at 0\%-20\% cloud cover, and the reconstruction results gradually deteriorate as the cloud cover increases to 80\%-100\%. For example, in the fourth column of Fig. \ref{fig9}, the results of the Pix2pix model's restoration are similar at different cloud cover levels, with significant changes in the cloud-free areas and substantial differences in color distribution compared to the ground truth. Other methods rely on cloud-free information on cloudy images, and as cloud cover increases, the available auxiliary information decreases, leading to a decrease in restoration accuracy. All models can reconstruct the geometric structure of buildings when cloud cover is low. As cloud cover increases, the SpAGAN model and the SAR-opt-cGAN model produces undesirable artifacts, while the DSen2-CR model, GLF-CR model, and USSRN-CR model generate blurry reconstruction results and fail to correctly reconstruct the texture details of buildings. The performance of our model also decreases as the cloud cover increased, it still achieves good visual effect. Even at high cloud cover levels of 80\%-100\%, the model can reconstruct the high-frequency texture details and structural information of buildings while maintaining edge consistency.

In Fig. \ref{fig10}, all models obtain higher PSNR scores for barren and sparse vegetation, cropland, and water categories. However, for complex textured features such as building and traffic route, the pixel-level PSNR scores are lower, indicating lower reconstruction accuracy. For spectral characteristics, all models demonstrate good spectral performance for barren and sparse vegetation and water categories but exhibit significant spectral deviations in the tree cover and grassland categories. Also, they achieve the highest reconstruction accuracy at 0\%-20\% cloud cover and the lowest performance at 80\%-100\% cloud cover.

Specifically, the Pix2pix model relies solely on PolSAR data for reconstruction, resulting in the lowest PSNR, SSIM, CC, and SAM scores across all categories, significantly lower than other algorithms. The performance of the Pix2pix model is independent of cloud cover, with PSNR values fluctuating around 23 as cloud cover increases. The SpAGAN model outperforms the Pix2pix model in reconstruction accuracy across all categories and cloud cover. Except for having PSNR scores similar to the SAR-opt-GAN model in the water categories, it lags behind in other categories compared to other models based on PolSAR and optical image. Therefore, multimodal restoration achieves better accuracy than single modal restoration. As the cloud cover increases, the DSen2-CR model, GLF-CR model, USSRN-CR model, and the proposed PODF-CR model mitigate the performance degradation caused by increasing cloud cover in terms of SSIM and CC metrics. The PODF-CR model fully considers the characteristics of PolSAR images, better integrates complementary information, and achieves better feature representation for optical image reconstruction through refinement operations. Therefore, the proposed method achieves the highest quantitative results at all cloud cover levels and all categories, and exhibits a significant improvement in the PSNR scores for building, traffic route, and water.

In summary, our model has significant advantages in visual effects, spectral restoration, and texture reconstruction, and it exhibits better quantitative metrics compared to other contrast methods.

\subsection{Further Discussion}

PODF-CR network can better integrate optical images with SAR images, refine the extracted features to obtain a better feature representation, and denoise SAR images and extract multi-scale feature information to provide better conditions for the subsequent decoding and restoration of the fused images, thereby improving cloud removal performance. In this section, we conducted experiments in both the input data and model aspects to demonstrate the effectiveness. Furthermore, this section includes parameter sensitivity analysis and computational cost.

\subsubsection{Ablation Study}
\subsubsubsection{Importance of Input Image}

\begin{figure*}[!t]
  \centering
  \includegraphics[width=\textwidth]{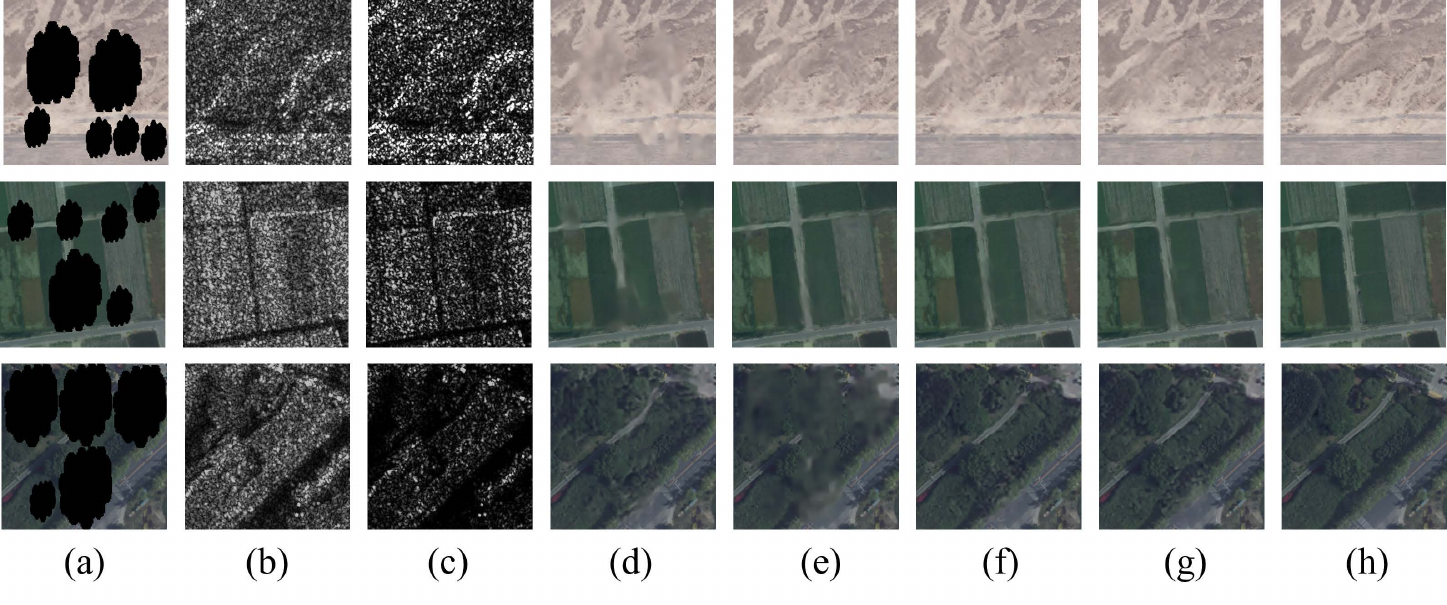}
  \caption{Ablation experiments based on input data of PODF-CR algorithm. (a) Cloudy images. (b) BCFSAR images. (c) PFSAR images. (d) Results of the w/o PolSAR model. (e) Results of the w PolSAR model. (f) Results of the w BCFSAR model. (g) Results of the w PFSAR model. (h) Ground truth.}
  \label{fig11}
\end{figure*}

In order to better evaluate the impact of various feature images in the dataset on the restoration results, we conducted ablation experiments on the dataset. In the experiments, the model that only uses BCFSAR images as input is denoted as the w BCFSAR model, the model that only uses PFSAR images is denoted as the w PFSAR model, and the model that uses both is denoted as the w PolSAR model. The model that uses only cloudy optical data as input, denoted as the w/o PolSAR model. The qualitative and quantitative evaluation results of the ablation experiments of these data are shown in Fig. \ref{fig11} and Table \ref{tab:table_4}.

Fig. \ref{fig11} displays the results of barren and sparse vegetation, cropland, and tree cover. Overall, the w PFSAR model exhibits more texture details and colors similar to the cloud-free optical images compared to the w PolSAR model and w BCFSAR model, particularly for cropland and barren and sparse vegetation. Therefore, in the proposed model, PFSAR feature images play a greater role compared to BCFSAR feature images, providing more information for image restoration. Additionally, the model was trained without PolSAR data, denoted as w/o PolSAR, to validate the importance of PoSAR images. As the input is a single-source signal, i.e., the optical image itself, a single-stream network is employed without a fusion strategy. Removing all auxiliary data, w/o PolSAR tends to produce oversmoothed results in cloud-covered areas and fails to reconstruct texture details in tree areas. This demonstrates the complementarity of PolSAR images to optical images. The quantitative results in Table \ref{tab:table_4} also indicate that the w PFSAR model achieves the highest performance in terms of PSNR and SAM metrics. When using only BCFSAR feature images or PoSAR images for restoration, the accuracy of the reconstructed image is reduced. Furthermore, removing the auxiliary PolSAR data leads to significant decreases in all metrics, with the PSNR metric decreasing by 2.2 dB.
Therefore, for the proposed model, PFSAR feature images provide more information than BCFSAR feature images. Compared with using both BCFSAR and PFSAR feature images as auxiliary inputs, the model achieves the highest accuracy when using only PFSAR feature images. Introducing BCFSAR feature images may introduce noise to the residual learning of the entire network, leading to a decrease in reconstruction accuracy. Additionally, when not using PolSAR data as auxiliary information, the overall model restoration accuracy significantly decreases, indicating that the proposed model effectively leverages PolSAR images for missing area information, enabling interaction between optical and PolSAR images.

Table 4 Quantitative ablation study of proposed algorithm without use of the PolSAR images, and with use of the PolSAR images, the backscatter coefficient feature images (BCFSAR), and the polarization feature SAR (PFSAR) images.

\begin{table*}[htbp]
  \centering
  \caption{Quantitative ablation study of proposed algorithm without use of the PolSAR images, and with use of the PolSAR images, the backscatter coefficient feature images (BCFSAR), and the polarization feature SAR (PFSAR) images. The best performance is shown in \textbf{bold}}
  \label{tab:table_4}
  \renewcommand{\arraystretch}{1.3}
  \begin{tabular}{ccccccc}
  \toprule
  Method & BCFSAR& PFSAR& PSNR$\uparrow$ & SSIM$\uparrow$ & CC$\downarrow$ & SAM$\downarrow$\\
  \midrule
w/o PolSAR &\ding{55}& \ding{55}& 32.772	&0.861	&0.877	&1.188 \\
w PolSAR	&\ding{51}& \ding{51}&34.874	&0.890	&0.921	&0.922\\
w BCFSAR	&\ding{51}& \ding{55}&34.101	&0.872	&0.907	&0.994\\
w PFSAR	&\ding{55}& \ding{51}&\textbf{34.992}	&\textbf{0.892}	&\textbf{0.922}	&\textbf{0.916}\\
  \bottomrule
  \end{tabular}
  \end{table*}

\subsubsubsection{Importance of the Module}

\begin{figure*}[!t]
  \centering
  \includegraphics[width=\textwidth]{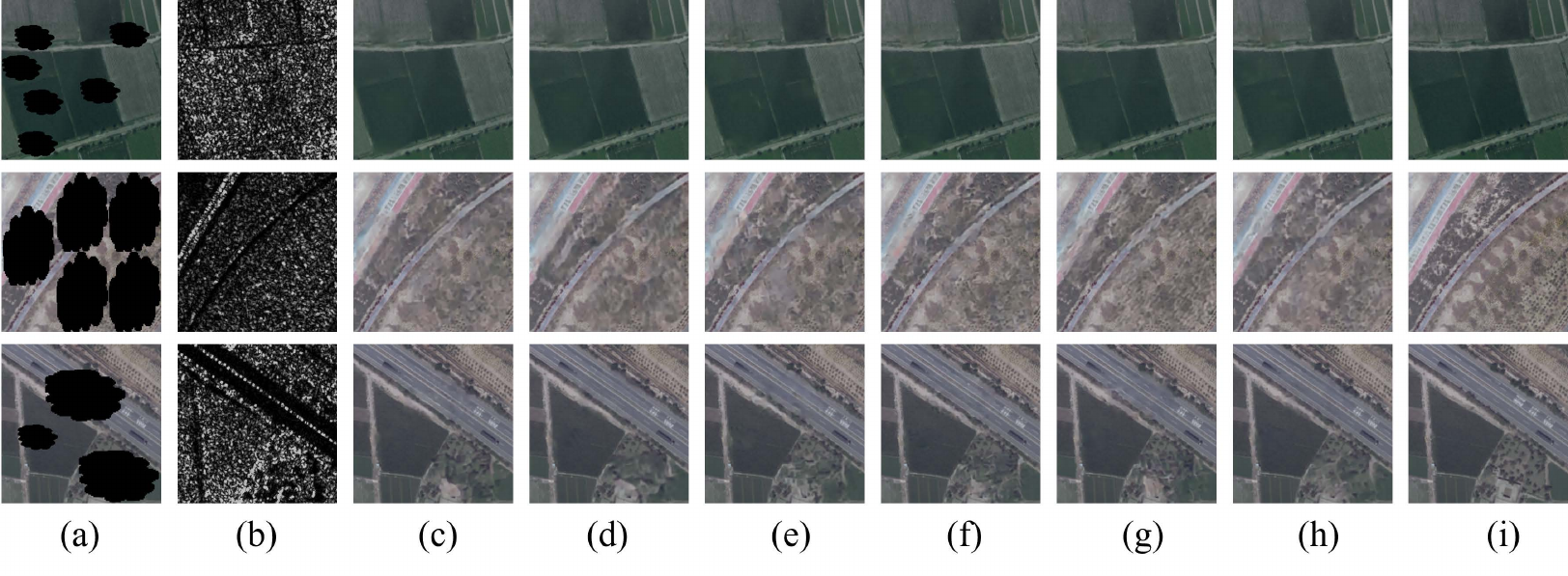}
  \caption{Ablation experiments based on the module of PODF-CR algorithm. (a) Cloudy images. (b) PFSAR images. (c) Results of the w/o SCDF model. (d) Results of the w/o GC model. (e) Results of the w/o MMCF model. (f) Results of the w/o MMRF model. (g) Results of the w/o ASPP model. (h) Results of the PODF-CR model. (i) Ground truth.}
  \label{fig12}
\end{figure*}

In terms of the model, we evaluated the effectiveness of various structures by comparing a few variants with and without use of the coupling spatial and channel dynamic filters (SCDF), gated convolution (GC), multi-modality cross fusion (MMCF), multi-modality refinement fusion (MMRF), and atrous spatial pyramid pooling (ASPP) to assess the importance of these operations. The qualitative and quantitative evaluation results of the ablation models and the proposed models are presented in Fig. \ref{fig12} and Table \ref{tab:table_5}.

1) Ablation Study of the SCDF: The SCDF was employed to address speckle noise in PolSAR images. To validate the effectiveness of the dynamic filter, we trained a network by removing the dynamic filter from the RB-DF block, denoted as w/o SCDF. The presence of speckle noise in SAR images significantly affects the restoration performance in cloudy areas. As evident from Table \ref{tab:table_5} and Fig. \ref{fig12}, the w/o SCDF model shows a decrease in all metrics and fails to reconstruct the correct texture of objects. The proposed SCDF alleviates the speckle noise issue in PolSAR images, resulting in clearer image generation.

2) Ablation Study of the GC: The GC can differentiate between valid and invalid pixels and effectively utilize information from cloud-free regions to extract features. The w/o GC model indicates that standard convolution replaced the gated convolution in our optical branch model. As shown in the second row of Fig. \ref{fig12}, the grassland reconstructed by w/o G-Conv model is fuzzy. Furthermore, the introduction of GC leads to improvements in all metrics, particularly with a 0.76 increase in PSNR. Qualitative and quantitative results demonstrate that GC can restore details and textures blurred by standard convolution.

3)Ablation Study of the MMCF: To study the role of the MMCF module in fusing multimodal PolSAR and optical data, we obtained fusion features by simply concatenating PolSAR and optical images, denoted as w/o MMCF. As shown in Table \ref{tab:table_5}, the introduction of the MMCF module leads to an improvement of 0.53 in PSNR. As depicted in the first row of Fig. \ref{fig12}, the w/o MMCF model yields artifacts for the cropland, while the proposed PODF-CR network based on the MMCF module achieves better fusion of PolSAR and optical image features, enabling the restoration of more complete texture structures and better visual effects.

4)Ablation Study of the MMRF: The MMRF module can suppress irrelevant information and emphasize complementary information, refining the fusion features of PolSAR and optical images from a global perspective to enhance the global consistency of the reconstructed images. By concatenating the fusion features obtained from PolSAR, optical, and the fusion features as inputs to the encoder, denoted as w/o MMRF, it is observed from Table \ref{tab:table_5} that the w/o MMRF model shows a decrease in PSNR, SSIM, CC, and SAM metrics. The MMRF module refines the fusion features from a global perspective, guiding the global interaction between optical and PolSAR features, effectively improving cloud removal performance and ensuring the predicted cloud-free images are more consistent. As shown in the second row of Fig. \ref{fig12}, the model with the introduced MMRF module can reconstruct the texture of the grassland with stronger detail preservation capabilities.

5) Ablation Study of the ASPP: Further validation of the effectiveness of the ASPP structure in utilizing spatial multiscale information was conducted by removing the ASPP module, denoted as w/o ASPP to train the PODF-CR network. As shown in the third row of Fig.\ref{fig12}, the "w/o ASPP" model also fails to reconstruct certain texture details of the traffic route, resulting in significant information loss. The qualitative results also indicate that the ASPP structure effectively utilizes multiscale information to improve cloud removal performance.

\begin{table}[htbp]
  \centering
  \caption{Quantitative ablation study of proposed algorithm with and without use of the coupling spatial and channel dynamic filters (SCDF), gated convolution (GC), multi-modality cross fusion (MMCF), multi-modality refinement fusion (MMRF), and atrous spatial pyramid pooling (ASPP). The best performance is shown in \textbf{bold}}
  \label{tab:table_5}
  \renewcommand{\arraystretch}{1.3}
  \begin{tabular}{ccccc}
  \toprule
  Method & PSNR$\uparrow$ & SSIM$\uparrow$ & CC$\downarrow$ & SAM$\downarrow$\\
  \midrule
w/o SCDF	&34.746	&0.887	&0.918	&0.954\\
w/o GC	&34.233	&0.873	&0.908	&0.970\\
w/o MMCF	&34.460	&0.878	&0.913	&0.961\\
w/o MMRF	&34.523	&0.878	&0.915	&0.974\\
w/o ASPP	&34.436	&0.877	&0.913	&0.952\\
PODF-CR	&\textbf{34.992}	&\textbf{0.892}	&\textbf{0.922}	&\textbf{0.916}\\
  \bottomrule
  \end{tabular}
  \end{table}

\subsubsection{Parameter Sensitive Analysis}

\begin{figure*}[!t]
  \centering
  \includegraphics[width=\textwidth]{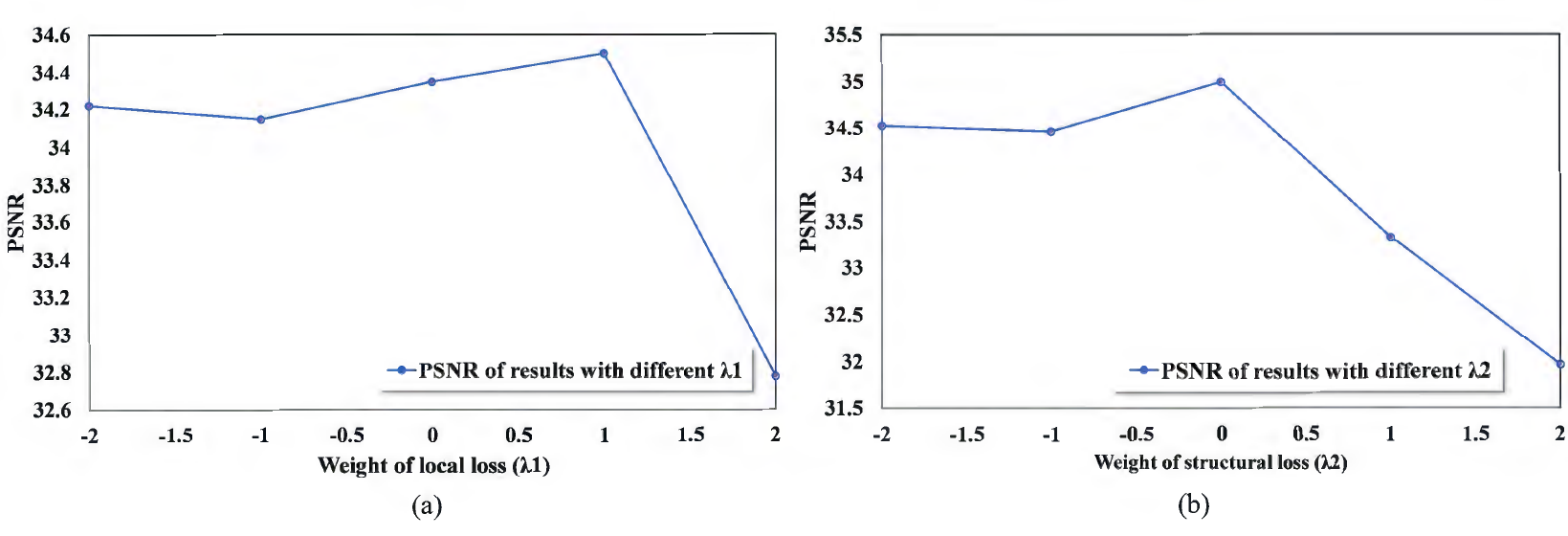}
  \caption{Results of the proposed model with different weight. (a) $\lambda1$. (b) $\lambda2$.}
   \label{fig13}
\end{figure*}

We mainly analyzed the weights $\lambda1$ and $\lambda2$ of the local loss and structural loss in the loss function in order to fine-tune the model to the optimal state. For quantitative evaluation, we compared models with different weights using PSNR. As shown in Fig. \ref{fig13}(a), the weight of $\lambda1$ varies between $10^{-2}$ and $10^2$, and it is observed that the PSNR reaches its maximum value when the weight of $\lambda1$ is fixed at 10. Then, by adding the structural loss, $\lambda2$ is determined as shown in Fig. \ref{fig13}(b). Finally, the best result is obtained when the value of $\lambda2$ is determined to be 1.

\subsubsection{Computational Complexity}

We present a comparison detailing the number of floating-point operations (FLOPs), model parameters, and the time in Table \ref{tab:table_6}. For a test set image of size 256 pixels × 256 pixels, as can be seen from FLOPs and parameter number, the PODF-CR model is relatively large, with a parameter number of 58.86, indicating that the model has a certain computational complexity. The time on the whole data set is 135s. In general, the reasoning time is faster than that of GLFCR model, which can achieve higher repair accuracy.

\begin{table}[htbp]
  \centering
  \caption{Comparison of the number of parameters, FLOPs, and time for various models.}
  \label{tab:table_6}
  \renewcommand{\arraystretch}{1.3}
  \begin{tabular}{cccc}
  \toprule
  Method & FLOPs(G) & Params(M) & Time(s)\\
  \midrule
SpAGAN & 17.21	&0.22	&51.56 \\
Pix2pix	&18.38	&54.41	&49.54\\
SAR-Opt-cGAN	&18.45	&54.42	&50.02\\
DSen2-CR	&1248.68	&18.92	&134.97\\
GLF-CR	&244.02	&14.8	&166.41 \\
USSRN-CR	&1233.25	&18.77	&113.50\\
PODF-CR	&418.84	&58.86	&135.02\\
  \bottomrule
  \end{tabular}
  \end{table}

\section{Conclusion}\label{Discussion and Conclusion}

This article proposes a cloud removal algorithm based on the fusion of PolSAR images and optical images, achieving the reconstruction of optical images. Specifically, to better facilitate image fusion, the PODF-CR model effectively realizes the interaction between multimodal optical images and PolSAR images based on cross-skip connections, and refines the extracted features using an attention mechanism to obtain better feature representation. Additionally, dynamic filters and multi-scale convolution are used to filter speckle noise in PolSAR images and obtain multi-scale information, thereby improving the restoration performance. Experimental results on the airborne dataset of PolSAR and optical data validate the feasibility and effectiveness of the model, as well as its superiority over other deep learning-based methods.

Although the proposed method has achieved satisfactory results in both qualitative and quantitative evaluations, there are still some limitations that need to be overcome. In future work, we will utilize multi-temporal PolSAR data and optical data for image restoration to make the model more robust and generalizable.

\section*{Declaration of competing interest}
The authors declare that they have no known competing financial interests or personal relationships that could have appeared to influence the work reported in this paper.

\section*{Acknowledgements}
This research was partially supported by the National Natural Science Foundation of China under Grant 42201503.

\printcredits

\bibliography{egbib}

\end{sloppypar}
\end{document}